\PassOptionsToPackage{table}{xcolor}
\documentclass{article} 
\usepackage{iclr2027_conference,times}

\usepackage{amsmath,amsfonts,bm}

\def\eqref#1{equation~\ref{#1}}

\def\1{\bm{1}}

\DeclareMathAlphabet{\mathsfit}{\encodingdefault}{\sfdefault}{m}{sl}
\SetMathAlphabet{\mathsfit}{bold}{\encodingdefault}{\sfdefault}{bx}{n}

\usepackage{graphicx}
\usepackage{microtype}
\usepackage{hyperref}
\usepackage{url}
\usepackage{enumitem}
\usepackage{booktabs}
\usepackage{multirow}
\usepackage[most]{tcolorbox}
\usepackage{wrapfig}
\usepackage{subfig}

\usepackage{adjustbox}
\usepackage{placeins}
\usepackage{array}
\usepackage{pifont}
\usepackage{fontawesome5}
\usepackage{etoc} 
\usepackage{wrapfig}
\definecolor{darkblue}{rgb}{0, 0, 0.5}
\hypersetup{colorlinks=true, citecolor=darkblue, linkcolor=darkblue, urlcolor=darkblue}

\newcommand{\hficon}{\raisebox{-0.15em}{\includegraphics[height=1.27em]{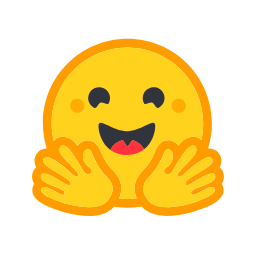}}}
\newcommand{\codelink}[1]{\href{#1}{\faGithub\ \textbf{Code}}}
\newcommand{\modellink}[1]{\href{#1}{\hficon\ \textbf{Models}}}

\title{Specializing Without Forgetting: Analyzing Knowledge Preservation in Multilingual Model Adaptation}

\author{Sanchit Ahuja \&  Terra Blevins \\
Northeastern University\\
\texttt{\{ahuja.sanc,t.blevins\}@northeastern.edu}
}

\newcommand{\preprintcopy}{\iclrfinalcopy}
\preprintcopy
\begin{document}

\maketitle
\lhead{Preprint}

\begin{center}
\small
\codelink{https://github.com/sanchit-ahuja/scaling-multilingual-experts} \quad\quad
\modellink{https://huggingface.co/collections/sanchitahuja205/x-elm-experts-gemma-3-4b}
\end{center}

\begin{abstract}

While continual pretraining (CPT) is a practical way to extend large language models to new languages, naïve finetuning often erodes existing capabilities through catastrophic forgetting. We investigate which model layers drive this trade-off, and whether interventions at these layers can guide knowledge preservation during adaptation. We interpolate \texttt{gemma-3-4b} model states before and after CPT on five language families to localize forgetting on reading comprehension and translation, finding that middle-layer reversion yields the largest comprehension recovery, while translation effects vary by language family and direction. Guided by these findings, we evaluate CPT strategies that leverage this layer information to mitigate forgetting: layer freezing, layer-range L2 regularization, post-hoc layer reversion,  and model souping, comparing all strategies against joint multilingual and family-specific vanilla CPT baselines. We find that preserving the layer weights identified via model interpolation substantially reduces comprehension loss relative to joint CPT, with layer freezing exceeding base model performance on average. However, these strategies yield mixed translation results: dense training or post-hoc reversion often outperforms both training-time constraints and family-specific specialization, complicating prior assumptions about \textit{how} models should be aligned when extended to new tasks. Instead, we argue that multilingual adaptation strategy should be informed by target language, base model knowledge, and downstream task, and propose interpolation-based localization as a diagnostic for identifying candidate layers before committing to a training-time intervention in a new setting.


\end{abstract}

\section{Introduction}

Adapting Large Language Models (LLMs) through continual pretraining (CPT) is a practical method of expanding model coverage to new languages while avoiding the prohibitive compute costs of pretraining from scratch~\citep{zhao2025babelopenmultilinguallarge,dou2024sailoropenlanguagemodels}. However, naïve CPT on target languages yields strong language acquisition but often causes catastrophic forgetting~\citep{MCCLOSKEY1989109, kirkpatrick2017overcoming} of the model's original knowledge; this is especially true in multilingual settings, where the curse of multilinguality~\citep{conneau-etal-2020-unsupervised} forces parametric trade-offs between language coverage and preserving existing capabilities. Furthermore, this balance cannot be assessed through language modeling alone, as lower target-language perplexity does not imply better downstream task performance~\citep{tay2022scaleefficientlyinsightspretraining,li-etal-2022-overcoming}. 

Mechanistic analysis of LLMs has found that middle layers of transformer LMs are associated with language-agnostic processing, while early and late layers handle language-specific inputs and outputs~\citep{wendler-etal-2024-llamas,NEURIPS2024_1bd359b3,dumas-etal-2025-separating}.
Recent work on cross-lingual reasoning exploits this separation during adaptation, showing that language and mathematical capabilities can be effectively combined by composing the outer layers from a language-adapted model with the middle layers from a reasoning-adapted model~\citep{bandarkar2025layerswappingzeroshotcrosslingual}, and that similar benefits can be obtained simply by freezing or reverting the middle layers during adaptation~\citep{bandarkar-peng-2025-unreasonable}. However, these findings come from supervised fine-tuning on a single task with limited CPT experiments, leaving open whether they extend to multilingual continual pretraining, cross-lingual evaluation, or generation tasks.

This paper generalizes this layer-aware preservation approach by first identifying \textit{which} layers specifically drive task knowledge change during joint multilingual (Dense) CPT with a new post-hoc interpolation diagnostic method (\S \ref{sec:analysis}), rather than relying on general insights from prior work. We then test whether preserving the layer parameters identified by this method improves knowledge acquisition and retention during CPT (\S \ref{sec:experiments}). Through these experiments, we test: (1) which layer updates drive comprehension loss and translation gains; (2) whether these effects are consistent across language families and tasks; and (3) whether layers identified through post-hoc intervention make effective training-time constraints. Specifically, we experiment with layer-aware preservation techniques (using the most impactful layers identified by our post-hoc analyses) for multilingual CPT of \texttt{gemma-3-4b}~\citep{gemmateam2025gemma3technicalreport} on 32 languages from five language families, comparing joint multilingual (Dense) CPT with independently trained family experts. We measure language acquisition with perplexity and downstream capabilities with reading comprehension and translation, on both CPT training languages (held-in) and related held-out languages.



\begin{figure}
    \centering
    \includegraphics[width=0.75\linewidth]{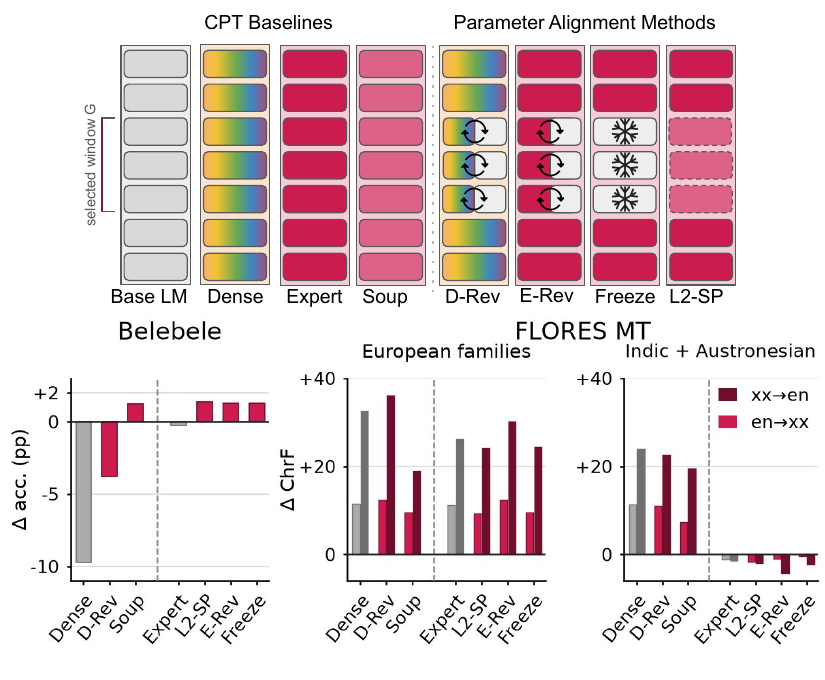}
      \caption{(a) CPT baselines and targeted parameter alignment methods. (b) Held-in changes relative to Base. D-Rev and E-Rev restore Base weights after CPT; Freeze holds the selected window $G$ fixed during CPT, and L2-SP regularizes it. The targeted methods use $G=[10,16)$ for Belebele and $G=[5,11)$ for FLORES. The FLORES panels show translation directions separately and average the named families within each group.}
    \label{fig:method}
\end{figure}

Our results show that the benefits of layer reversion are \textbf{strongly} task-dependent (Figure~\ref{fig:method}). In our post-hoc analyses, middle-layer reversion produces the largest comprehension recovery, corroborating prior work, but harms translation: restoring layers $[10,16)$ of the Dense model to Base improves reading comprehension accuracy by $5.9$ points but reduces translation by $3.5$ ChrF.\footnote{In contrast, restoring late layers $[27,33)$ reduces translation by $18.5$ ChrF while barely affecting comprehension accuracy ($0.4$).} We also find that layer-wise improvements identified by our interpolation analysis transfer to CPT for comprehension, as freezing, L2 regularization, and post-hoc reversion of layers $[10,16)$ all exceed unregularized family experts, Dense CPT, and the Base model on held-in reading comprehension, while also obtaining lower in-language perplexity than Base.

However, unlike prior work, the trends we observe for translation and held-out languages are less consistent: for translation, layer-preserving training paradigms with the identified layers [5,11) do not cleanly transfer to better results. Instead, post-hoc reversion of these layers improves over the respective baselines for both the Dense model and most family experts, whereas freezing or regularizing the same window fails to improve over baseline in three of five families. Moreover, dense training on multiple languages often outperform specializing to the target languages, particularly for the lower-resourced Indic and Austronesian families. For held-out languages, we similarly see that while targeted layer preservation improves comprehension over vanilla experts, no intervention approach surpasses the Base model; our translation gains also only hold when translating held-out languages to English (\textsc{xx}$\to$\textsc{en}: $31.8\to51.3$ ChrF under Dense CPT, versus $34.4\to34.6$ for \textsc{en}$\to$\textsc{xx}). 

These results indicate that the layer regions that store downstream knowledge vary across tasks and that transferring layers to preserve across tasks can significantly harm performance. Furthermore, while the regions identified by our post-hoc interpolation analysis can inform better, layer-aware CPT strategies, the benefits only seem to hold if downstream task knowledge is otherwise lost during targeted CPT (e.g., on comprehension). We also note that language family specialization is not always the best method for improving target language task performance, contrary to prior assumptions. Thus, rather than a single preservation strategy for improving CPT, we argue that multilingual adaptation should be informed by the target task, base model knowledge, and languages, with interpolation-based localization used to identify candidate layers \textit{before} committing to a training-time intervention.

The primary contributions of this work are:
\begin{itemize}
    \item A new diagnostic method that uses layer interpolation to systematically identify which layer ranges drive performance gains and losses during model adaptation, along with a corresponding analysis demonstrating that the layers affecting performance changes after multilingual CPT can vary significantly across task types and target languages (\S\ref{sec:analysis}).
    \item A controlled comparison of training-time parameter preservation strategies (layer freezing, layer-range L2-SP) against post-hoc reversion and souping, applied at the identified windows by our interpolation analysis across five language families, showing that localization can transfer training-time preservation for comprehension but not for translation (\S\ref{sec:experiments}). 
    \item An examination of the acquisition--forgetting trade-off after CPT on both held-in and held-out languages, identifying the limits of improving language-specific behavior while preserving downstream performance from the base model (\S\ref{sec:heldout}). 
\end{itemize}

\section{Experimental Setting}


In this paper, we investigate how to leverage layer-specific parametric knowledge to mitigate catastrophic forgetting while retaining the benefits of language-specific adaptation during multilingual continual pretraining. We first use layer-wise interventions to characterize the downstream effects of CPT updates (\S \ref{sec:analysis}), then test whether these findings can guide targeted preservation during training (\S \ref{sec:experiments}). In this section, we detail our overall experimental setup, which performs joint multilingual and family-specific CPT from a shared base (\texttt{gemma-3-4b}) checkpoint with various parameter-alignment strategies(Figure~\ref{fig:method}), and evaluates the resulting models on reading comprehension, translation, and in-domain perplexity.

\subsection{Model, Data and Language Coverage}
\label{sec:families}

All our experiments build on the text-only pre-trained version of \texttt{gemma-3-4b}~\citep{gemmateam2025gemma3technicalreport}, which we refer to as the base checkpoint. We construct a dataset for continued pretraining from MADLAD-400~\citep{kudugunta2023madlad}, covering 32 languages across the Slavic, Germanic, Indic, Austronesian, and Romance language families; we additionally evaluate model generalization on 22 held-out languages across these families (Table~\ref{tab:language_families}).\footnote{We note that, since the Gemma 3 training language distribution is not publicly reported, we are unable to verify whether the held-out languages were seen during the original training process.} For each family, we sample 5B token corpus split uniformly across languages; Dense CPT trains on the union of the five training corpora (25B tokens). Documents are tokenized with the Gemma-3 tokenizer at a maximum sequence length of 2,048 tokens, using a 95\%/5\% train/validation split.

\subsection{Dense, Family-Specific CPT and Souping}
\label{sec:baselines}
We consider three unregularized CPT settings initialized from the same Base checkpoint :

\textbf{Dense CPT}\quad trains a single model jointly on the combined corpus spanning all 32 training languages.

\textbf{Family Expert}\quad trains a separate model on each language family initialized from the Base model.

\textbf{Expert Soup}\quad After training five family experts, we merge them into a single model by \textbf{uniformly averaging} their weights: $\theta_{\text{soup}} = \frac{1}{5}\sum_{f=1}^{5} \theta_f$, where $\theta_f$ are the weights of family $f$'s expert.

The Dense checkpoint provides the starting point for the layer-interpolation analysis in \S\ref{sec:analysis}. Family-specific checkpoints allow us to examine whether the identified intervention effects transfer across adaptation settings. Souping tests whether uniformly averaging independently trained family experts can recover base capabilities without any training-time constraint.

\begin{table*}[t]
\footnotesize
\centering
\fontsize{8}{8}\selectfont
\begin{tabular}{p{1.4cm} p{6.6cm} p{4.4cm}}
\toprule
\textbf{Family} & \textbf{Training Languages} & \textbf{Held-Out Languages} \\
\midrule
Slavic       & Macedonian (mk), Croatian (hr), Russian (ru), Slovak (sk), Serbian (sr), Ukrainian (uk) & Bulgarian (bg), Czech (cs), Polish (pl), Slovenian (sl), Lithuanian (lt), Latvian (lv) \\[1pt]
\midrule
Germanic     & Afrikaans (af), Frisian (fy), Luxembourgish (lb), Danish (da), Dutch (nl), English (en) & Swedish (sv), Norwegian (no), Icelandic (is) \\[1pt]
\midrule
Indic        & Bengali (bn), Hindi (hi), Kannada (kn), Malayalam (ml), Marathi (mr), Nepali (ne), Tamil (ta), Telugu (te) & Assamese (as), Gujarati (gu), Odia (or), Punjabi (pa), Sinhala (si), Urdu (ur), Sindhi (sd) \\[1pt]
\midrule
Austronesian & Samoan (sm), Javanese (jv), Cebuano (ceb), Filipino (fil), Indonesian (id), Malay (ms) & Ilocano (ilo), Waray (war), Malagasy (mg), M\={a}ori (mi), Sundanese (su) \\[1pt]
\midrule
Romance      & Spanish (es), Portuguese (pt), French (fr), Galician (gl), Italian (it), Romanian (ro) & Catalan (ca) \\
\bottomrule
\end{tabular}
\caption{Language families, their training languages, and held-out languages for evaluating within-family generalization.}
\label{tab:language_families}
\end{table*}

\subsection{Training and Evaluation}
\label{sec:setup}
\textbf{Training} \quad CPT uses bfloat16 precision and gradient checkpointing. The Dense and family-specific runs use different training schedules and stopping criteria, detailed in Appendix~\ref{app:training_details}. The schedules and comparison controls for the targeted preservation experiments are specified in \S\ref{sec:experiments}.

  \textbf{Evaluation} \quad We measure reading comprehension using Belebele
  ~\citep{bandarkar-etal-2024-belebele}, translation using FLORES-200 ChrF in both translation directions~\citep{nllbteam2022languageleftbehindscaling}, and language acquisition using perplexity on held-out MADLAD text. Belebele and FLORES evaluations use \texttt{lm-eval-harness} in a 2-shot setting~\citep{eval-harness}.
  FLORES scores follow a shared first-line post-truncation protocol across checkpoints (Appendix~\ref{app:flores_eval_protocol}).

\section{Understanding Layer-Aware Adaptation}
\label{sec:analysis}
An open question for model adaptation is which regions mediate knowledge retention, under the hypothesis that selectively preserving these parameter values during CPT could mitigate catastrophic forgetting. Prior work suggests that the middle layers of transformer LMs support semantic and reasoning computations \citep{bandarkar-peng-2025-unreasonable,bandarkar2025layerswappingzeroshotcrosslingual,wendler-etal-2024-llamas}, while the initial and final layers encode language-specific information~\citep{tang2024languagespecificneuronskeymultilingual,kojima2024multilingualabilitydecoderbasedpretrained,NEURIPS2024_1bd359b3}. In this section, we test whether these layer intuitions hold for downstream reading comprehension and translation across language families. Specifically, we use interpolation analysis to localize which layer regions lead to the largest task performance changes after continual pretraining by interpolating between the original base and updated parameters within the target layers (Figures~\ref{fig:causal_belebele_layer_sweep} and~\ref{fig:causal_flores_layer_sweep_family}).

\subsection{Causal Analysis of Layer Drift}
\label{sec:causal_analysis}

 We perform a targeted interpolation analysis on the unregularized Dense CPT model. For a layer group $G$, we replace that group's parameters with $\theta_G(\alpha)=\theta_G^0+\alpha(\theta_G^{\text{Dense}}-\theta_G^0)$, while keeping all other layers fixed to the Dense CPT checkpoint. $\alpha{=}0$ reverts the selected layer group to the base model and $\alpha{=}1$ recovers the original Dense CPT parameters for that group. We sweep $\alpha\in\{0,0.25,0.5,0.75,1\}$ over the first ($9$), middle ($19$), and last ($6$) layer groups  \footnote{Motivated by prior outer/middle layer decompositions~\citep{bandarkar-peng-2025-unreasonable}, we adopt a coarse 9/19/6 partition.} 
  and evaluate held-in language Belebele accuracy and ChrF for FLORES. These interventions measure the effect of replacing a selected group's CPT updates while holding the rest of the remaining weights fixed.

\begin{wrapfigure}{r}{0.45\linewidth}
\vspace{-\intextsep}
\centering
\includegraphics[width=\linewidth]{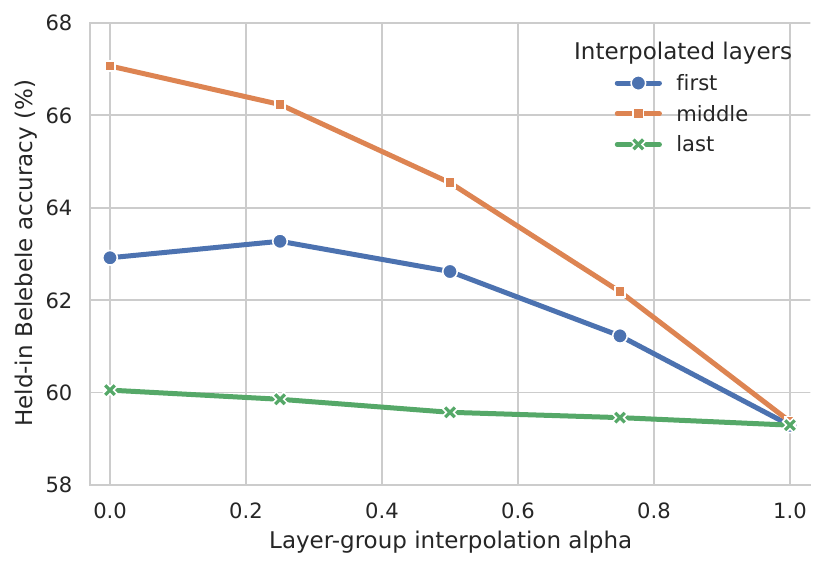}
\caption{Layer interpolation between the base model and Dense CPT for held-in Belebele accuracy. All non-interpolated layers are kept at their Dense CPT values.}
\label{fig:causal_belebele_layer_sweep}
\vspace{-\intextsep}
\end{wrapfigure}

Figure~\ref{fig:causal_belebele_layer_sweep} shows that retaining more of Dense CPT updates in the middle layer produces a monotonic $7.69$ point drop in Belebele accuracy. Middle-layer reversion therefore produces the largest recovery among these interventions. (Appendix~\ref{app:causal_layer_sweep} has family-wise results)

\begin{figure}[h]
\centering
\includegraphics[width=\textwidth]{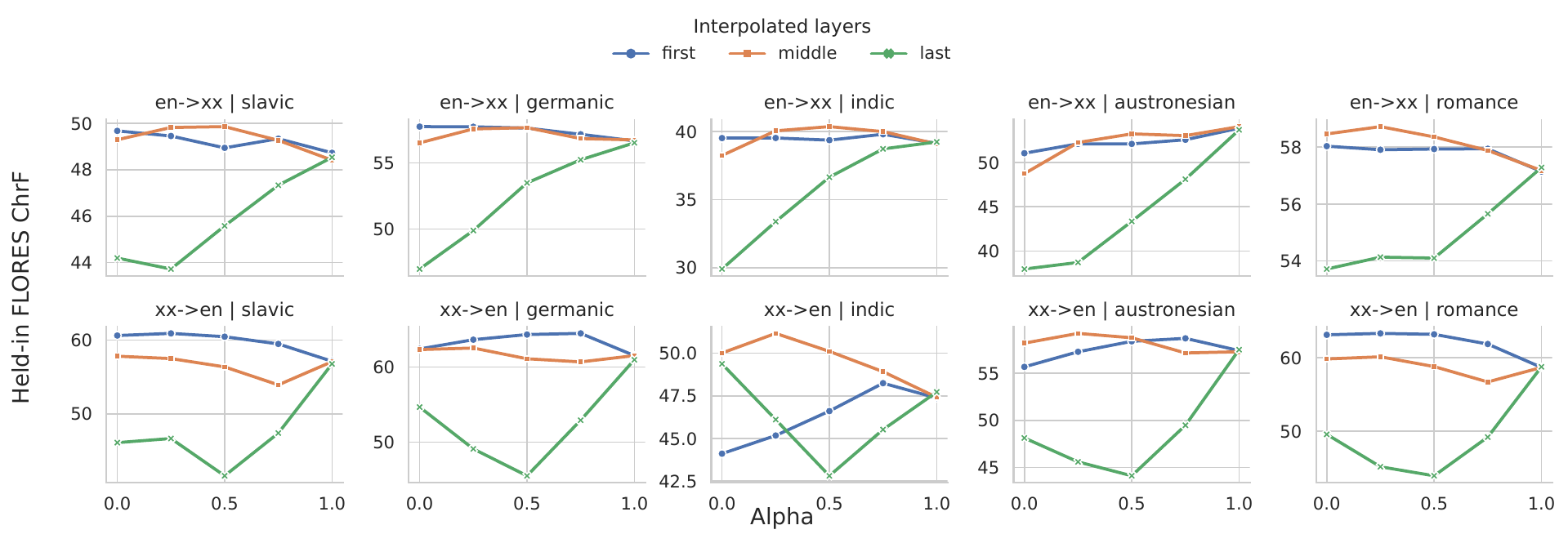}
\caption{Held-in FLORES ChrF under first-, middle-, and last-layer interpolation, broken down by language family (columns) and translation direction (rows). Scores use the same first-line post-truncation re-scoring protocol as the main FLORES tables.}
\label{fig:causal_flores_layer_sweep_family}
\end{figure}

FLORES exhibits a different layer-sensitivity profile, with effects varying across language families and translation direction (Figure~\ref{fig:causal_flores_layer_sweep_family}). The numbers are reported relative to the corresponding Dense endpoint i.e. $\Delta\mathrm{ChrF}=\mathrm{ChrF}(\alpha=0)-\mathrm{ChrF}(\alpha=1)$. Reverting the last layer group reduces EN$\to$\textsc{xx} performance in every family, with losses ranging from $3.57$ to $15.77$ ChrF. For \textsc{xx}$\to$EN, the same intervention harms four of the five families; Indic is the exception, improving by $1.65$ ChrF. These translation effects contrast with the small Belebele change under last-group reversion.

The effects of first-group reversion are more mixed. For \textsc{xx}$\to$EN, restoring base model weights improve Slavic, Germanic, and Romance by $3.49$, $0.89$, and $4.44$ ChrF respectively, but reduces Indic and Austronesian performance by $3.27$ and $1.72$ ChrF. The corresponding EN$\to$\textsc{xx} gains for Slavic, Germanic, and Romance are $0.95$, $1.08$, and $0.89$ ChrF. Thus, the benefit is more pronounced into English for Slavic and Romance, while Germanic shows small gains in both directions (Figure \ref{fig:causal_flores_layer_sweep_family}).

Middle-group reversion also produces these mixed results. While the gains are positive across all families, for \textsc{xx}$\to$EN direction ranging from $0.75$ to $2.60$ ChrF, the same is not true for  EN$\to$\textsc{xx}.


\subsection{Localizing task-dependent Reversion Effects}
\begin{wraptable}{r}{0.46\linewidth}
\vspace{-\intextsep}
\centering
\footnotesize
\setlength{\tabcolsep}{2pt}
\begin{tabular}{@{}lrr@{}}
\toprule
\shortstack{\textbf{Layers} \\ \textbf{reverted}} & \shortstack{\textbf{Belebele}\\$\Delta$ accuracy (pp)} & \shortstack{\textbf{FLORES}\\$\Delta$ ChrF} \\
\midrule
$[0,6)$ & $+2.2$ & $+0.26$ \\
$[5,11)$ & $+4.1$ & $\mathbf{+0.75}$ \\
$[10,16)$ & $\mathbf{+5.9}$ & $-3.47$ \\
$[24,30)$ & $-0.3$ & $-4.28$ \\
$[27,33)$ & $+0.4$ & $-18.50$ \\
$[28,34)$ & $+0.7$ & $-8.00$ \\
\bottomrule
\end{tabular}
\caption{Selected six-layer reversions to Base weights: changes relative to Dense CPT on Belebele (28 languages) and FLORES (20-direction screen). Positive values indicate improvement. Family-wise results in Appendix~\ref{app:causal_layer_sweep}.}
\label{tab:layer_window_belebele}
\vspace{-\intextsep}
\end{wraptable}

The three partitions that we studied above differ in width and may have wide variation within each group. We therefore refine this analysis using a fixed-width sliding window reversion, which holds the number of restored layers constant while varying their position. For each six-layer window $G_s = [s,s+6)$, we restore the selected parameters to their Base-model values, corresponding to $\alpha=0$, while retaining CPT weights everywhere else. This experiment measures the effect of the same-sized intervention at different depths and tests whether the coarse layer-group findings reflect localized sensitivity. This analysis helped us in identifying the highest-scoring window among those evaluated for both Belebele and FLORES (Table~\ref{tab:layer_window_belebele}).


Reverting $[10,16)$ produces the largest Belebele recovery, improving accuracy by $5.9$ pp over Dense CPT, but reduces FLORES performance by $3.47$ ChrF. In contrast, reverting the earlier window $[5,11)$ improves Belebele by $4.1$ points and yields a $0.75$ ChrF gain on the FLORES screening set (Appendix \ref{app:causal_layer_sweep} has language family wise gains). Late-layer reversion shows another contrast: restoring $[27,33)$ reduces FLORES performance by $18.50$ ChrF, while changing Belebele accuracy by only $+0.4$ points. A region whose reversion benefits comprehension need not benefit translation, and translation can be highly sensitive to interventions with little effect on comprehension.

We select $[10,16)$ for Belebele and $[5,11)$ for FLORES to perform CPT with freezing and Layer Range L2-SP applied to these layers on the base checkpoint to check how our findings transfer to these layer-specific training methodologies.

\section{Testing Targeted Interventions}
\label{sec:experiments}

\subsection{Targeted Preservation Strategies}
\label{sec:strategies}


For a selected window $G$, we compare two training-time preservation strategies, both initialized from Base parameter $\theta^0$.

Figure~\ref{fig:method} summarizes our alignment strategies and baselines (\S\ref{sec:baselines}):

\textbf{Layer Freezing}\quad We hold all parameters within $G$ fixed at their base values throughout CPT, while allowing parameters outside $G$ to update. This imposes a hard preservation constraint on the selected region.

\textbf{Layer-Range L2-SP}\quad We allow all parameters to update but penalize deviations from Base within $G$ using L2 starting-point regularization~\citep{li2018explicit} as adapted by \citet{kumar2024maintainingplasticitycontinuallearning}, with layer-dependent penalty strengths, offering a soft alternative to layer freezing during training. :$\mathcal{L}(\theta) = \mathcal{L}_{\mathrm{CPT}}(\theta) + \lambda \sum_{l \in G} \|\theta_l-\theta_l^0\|_2^2$. For the targeted experiments, we use $\lambda=0.05$ within the selected window.

  \textbf{Post-hoc Reversion}\quad We restore the same window to Base weights after unregularized CPT, retaining the adapted parameters elsewhere. This allows us to compare preservation during training with correction after training.


\subsection{Language Acquisition}
\label{sec:acq}

Table~\ref{tab:perplexity_main} shows the held-in validation perplexity for the two targeted windows across all families. The Dense CPT checkpoint had the best language acquisition compared to all our baselines ($7.20$) followed by the family experts ($7.30$). The targeted intervention experiments achieve similar average perplexities ($7.56$ - $7.62$) while still remaining below the base. This indicates that choosing either targeted window preserves broadly comparable in-family acquisition quality. Per-language breakdowns are in Appendix~\ref{app:perlang}. 

\begin{table*}[t]
\centering
\footnotesize
\setlength{\tabcolsep}{2pt}
\begin{tabular}{l rrrr rrrr rrrr}
\toprule
& \multicolumn{4}{c}{\textbf{Shared baselines}} & \multicolumn{4}{c}{$\mathbf{[5,11)}$} & \multicolumn{4}{c}{$\mathbf{[10,16)}$} \\
\cmidrule(lr){2-5} \cmidrule(lr){6-9} \cmidrule(lr){10-13}
\textbf{Family} & \textbf{Base} & \textbf{Dense} & \textbf{Expert} & \textbf{Soup} & \textbf{D.-Rev.} & \textbf{E.-Rev.} & \textbf{Freeze} & \textbf{L2-SP} & \textbf{D.-Rev.} & \textbf{E.-Rev.} & \textbf{Freeze} & \textbf{L2-SP} \\
\midrule
Slavic & 7.66 & 6.41 & \textbf{6.40} & 7.18 & 6.50 & 6.50 & 6.48 & 6.54 & 6.57 & 6.57 & 6.48 & 6.55 \\
Germanic & 11.41 & \textbf{7.18} & 7.23 & 9.90 & 7.50 & 7.53 & 7.50 & 7.54 & 7.49 & 7.53 & 7.49 & 7.49 \\
Indic & 10.07 & \textbf{7.66} & 7.73 & 8.83 & 8.02 & 8.04 & 8.04 & 8.09 & 7.98 & 8.04 & 8.00 & 7.98 \\
Austronesian & 11.78 & \textbf{6.22} & 6.74 & 10.10 & 7.31 & 7.31 & 7.29 & 7.28 & 6.91 & 7.27 & 7.20 & 7.20 \\
Romance & 10.01 & 8.55 & \textbf{8.40} & 9.28 & 8.63 & 8.67 & 8.65 & 8.64 & 8.82 & 8.67 & 8.63 & 8.68 \\
\midrule
Average & 10.19 & \textbf{7.20} & 7.30 & 9.06 & 7.59 & 7.61 & 7.59 & 7.62 & 7.55 & 7.62 & 7.56 & 7.58 \\
\bottomrule
\end{tabular}
\caption{Held-in validation perplexity $\downarrow$, averaged over training languages within each family. D.-Rev. restores Dense layers in the indicated window to Base; Freeze and L2-SP preserve the indicated layers during family CPT, while E.-Rev. restores the indicated layers to Base after Expert CPT. \textbf{Bold} = lowest value within a row.}
\label{tab:perplexity_main}
\end{table*}



\subsection{Preserving Reading Comprehension}
\label{sec:forget}

 We next evaluate whether targeted interventions preserve reading comprehension after CPT. Table~\ref{tab:belebele_main} summarizes Belebele results across language families and strategies. Dense CPT causes substantial forgetting, with accuracy dropping $6.6$--$12.3$ percentage points relative to Base across families, whereas vanilla family experts largely retain Base performance. All three targeted interventions outperform vanilla experts in every family. Together with the perplexity results, these comparisons illustrate the acquisition--forgetting trade-off: targeted interventions achieve higher reading-comprehension accuracy but slightly higher in-domain perplexity than vanilla family experts.
\begin{table*}[t]
\centering
\footnotesize
\setlength{\tabcolsep}{2pt}
\begin{tabular}{l rrrr rrrr}
\toprule
\textbf{Family} & \textbf{Base} & \textbf{Dense} & \textbf{Expert} & \textbf{Soup} & \textbf{D.-Rev.} & \textbf{E.-Rev.} & \textbf{Freeze} & \textbf{L2-SP} \\
\midrule
Slavic & 0.719 & 0.619 & 0.726 & \textbf{0.734} & 0.675 & \textbf{0.734} & 0.727 & 0.731 \\
Germanic & 0.758 & 0.642 & 0.756 & 0.761 & 0.708 & 0.759 & 0.762 & \textbf{0.763} \\
Indic & 0.601 & 0.535 & 0.593 & \textbf{0.629} & 0.584 & 0.624 & 0.623 & 0.627 \\
Austronesian & 0.668 & 0.587 & 0.661 & 0.682 & 0.642 & 0.688 & \textbf{0.698} & 0.692 \\
Romance & 0.748 & 0.625 & 0.746 & 0.749 & 0.696 & \textbf{0.753} & 0.748 & 0.750 \\
\midrule
Average & \shortstack{0.699\\[-1pt]{\scriptsize [0.693, 0.705]}} & \shortstack{0.602\\[-1pt]{\scriptsize [0.595, 0.608]}} & \shortstack{0.696\\[-1pt]{\scriptsize [0.691, 0.702]}} & \shortstack{0.711\\[-1pt]{\scriptsize [0.706, 0.717]}} & \shortstack{0.661\\[-1pt]{\scriptsize [0.655, 0.667]}} & \shortstack{0.712\\[-1pt]{\scriptsize [0.706, 0.717]}} & \shortstack{0.711\\[-1pt]{\scriptsize [0.706, 0.717]}} & \shortstack{\textbf{0.713}\\[-1pt]{\scriptsize [0.707, 0.718]}} \\
\bottomrule
\end{tabular}
\caption{Held-in Belebele accuracy $\uparrow$ (2-shot), averaged over training languages within each family. D.-Rev., E.-Rev., Freeze, and L2-SP target $[10,16)$; Base, Dense, Expert, and Soup are archival baselines. Average intervals are 95\% bootstrap CIs, macro-averaged over languages within each family and then over families. \textbf{Bold} = best per row.}
\label{tab:belebele_main}
\end{table*}


\subsection{Translation Quality (FLORES-200)}
\label{sec:flores}


Table~\ref{tab:flores_targeted_training} reports results for interventions targeting $[5,11)$\footnote{Appendix~\ref{app:flores_eval_protocol} describes the scoring protocol, and Appendix~\ref{app:flores_targeted} gives direction-wise results and comparison limitations.}. Dense reversion has family-dependent effects: restoring $[5,11)$ layers improves the Dense family-macro average from $53.7$ to $54.7$ ChrF. Slavic, Germanic, Austronesian, and Romance improve, whereas Indic declines.

For family experts, post-hoc reversion improves translation in four families, but the training-time preservation yields mixed results. Both freezing and L2-SP outperform vanilla experts on Germanic and Indic, but score less on Slavic, Austronesian and Romance. The consistent gains that we observed earlier with Belebele, do not cleanly transfer here to FLORES. One possible explanation is that anchoring layers to Base constrains updates useful for translation. Also, freezing and L2-SP constrain the selected layers during training and that might change how the rest of the layers adapt, whereas post-hoc reversion retains their unconstrained updates.

We also noticed a huge discrepancy with Austronesian results. While the Dense, D.-Rev. and Expert Soup outperform Base, the vanilla family expert and its targeted variants score below Base. Inspection of their outputs shows a high empty-rate (Appendix~\ref{app:flores_targeted}) for these checkpoints. Motivated by the late-layer sensitivity observed in the Dense sliding-window analysis, we investigated whether restoring $[27,33)$ to Base weights mitigate this failure in the Austronesian expert. The reverted checkpoint achieves $40.3$ ChrF with a $7.0\%$ empty-output rate, compared with $25.0$ ChrF and $57.5\%$ for the vanilla Austronesian expert in Table~\ref{tab:flores_targeted_training} ($29.7$ ChrF and $38.9\%$ in a separate generation run of the same expert). This improvement also leads to an increase in held-in perplexity from $6.75$ to $7.95$, illustrating a trade-off between translation recovery and in-domain language modeling. These results suggest that a localized intervention can mitigate the generation failure, but do not establish its underlying cause (Appendix ~\ref{app:flores_targeted}).
\begin{table*}[t]
\centering
\setlength{\tabcolsep}{3pt}
\footnotesize
\begin{tabular}{l rrrr rrrr}
\toprule
\textbf{Family} & \textbf{Base} & \textbf{Dense} & \textbf{Expert} & \textbf{Soup} & \textbf{D.-Rev.} & \textbf{E.-Rev.} & \textbf{Freeze} & \textbf{L2-SP} \\
\midrule
Slavic & 33.6 & 52.7 & 47.4 & 46.2 & \textbf{54.7} & 51.3 & 47.1 & 46.9 \\
Germanic & 34.8 & 59.0 & 57.0 & 50.4 & \textbf{61.3} & 59.7 & 57.7 & 58.1 \\
Indic & 29.9 & \textbf{43.4} & 36.1 & 42.6 & 41.3 & 36.4 & 36.3 & 36.6 \\
Austronesian & 33.8 & 55.5 & 25.0 & 48.1 & \textbf{56.2} & 22.2 & 24.7 & 23.4 \\
Romance & 34.4 & 57.9 & 54.9 & 49.2 & \textbf{60.2} & 56.4 & 49.5 & 48.5 \\
\midrule
Average & \shortstack{33.3\\[-1pt]{\scriptsize [33.1, 33.5]}} & \shortstack{53.7\\[-1pt]{\scriptsize [53.5, 54.0]}} & \shortstack{44.1\\[-1pt]{\scriptsize [43.8, 44.4]}} & \shortstack{47.3\\[-1pt]{\scriptsize [47.0, 47.6]}} & \shortstack{\textbf{54.7}\\[-1pt]{\scriptsize [54.6, 54.9]}} & \shortstack{45.2\\[-1pt]{\scriptsize [44.9, 45.5]}} & \shortstack{43.1\\[-1pt]{\scriptsize [42.8, 43.4]}} & \shortstack{42.7\\[-1pt]{\scriptsize [42.4, 43.0]}} \\
\bottomrule
\end{tabular}
\caption{Held-in FLORES-200 ChrF $\uparrow$ (2-shot), averaged over languages and both directions within each family; Average weights families equally. D.-Rev., E.-Rev., Freeze, and L2-SP\ target $[5,11)$. All Average cells include percentile 95\% bootstrap CIs. \textbf{Bold} = best per row; See Appendix~\ref{app:flores_targeted} for scoring and comparison details.}
\label{tab:flores_targeted_training}
\end{table*}

\subsection{Out-Of-Domain Generalization}
\label{sec:heldout}

\begin{figure}[h]
\centering
\includegraphics[width=0.8\textwidth]{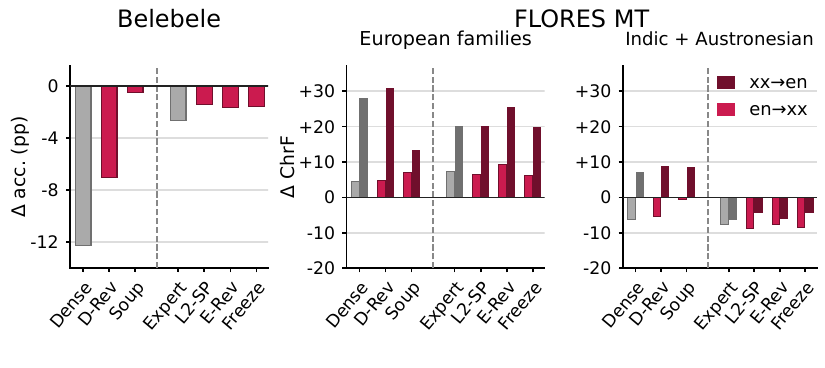}
\caption{Held-out Belebele accuracy delta and FLORES MT ChrF delta relative to the base model.}
\label{fig:heldout_results}
\end{figure}

We next ask whether the acquisition--forgetting patterns observed on CPT training languages extend to held-out languages within the same family groups. We evaluate 22 languages excluded from CPT (Appendix~\ref{tab:language_families}).\footnote{These languages are held out from our CPT data, not necessarily from Gemma's original pretraining.}

\textbf{Language modeling }\quad
Dense CPT increases held-out perplexity relative to Base in every family, and family-expert variants generally remain above Base. Expert Soup lowers perplexity in four of five families (Table~\ref{tab:perplexity_heldout}). Thus, the lower perplexity achieved on CPT training languages does not imply improved language modeling on their held-out relatives.

\textbf{Reading Comprehension}\quad
Freeze, L2-SP, and E.-Rev. improve held-out Belebele accuracy by approximately 1pp over vanilla experts, but remain below Base in every family, unlike their above-Base held-in averages. Expert Soup comes closest to Base overall (64.7\% versus 65.2\%. (Table~\ref{tab:belebele_targeted_heldout}, Figure~\ref{fig:heldout_results}).

\textbf{Translation}\quad
Across families and both directions, E.-Rev.\ improves the Expert reference from $38.4$ to $40.8$ ChrF, whereas Freeze and L2-SP both average $38.3$ (Tables~\ref{tab:heldout_flores_xx_en_perlang} and~\ref{tab:heldout_flores_en_xx_perlang}). Dense reversion likewise improves the average from $42.9$ to $44.4$ ChrF. Transfer is asymmetric: Dense improves xx$\to$en from $31.8$ to $51.3$ ChrF, but barely changes en$\to$xx ($34.4$ to $34.6$). Vanilla experts and their targeted variants also remain below Base on Indic and Austronesian. Together, these results reinforce that comprehension preservation does not ensure translation improvements, and that beneficial post-hoc reversion need not translate into effective training-time constraints.
\section{Related Work}



\textbf{Multilingual Catastrophic Forgetting}\quad Catastrophic forgetting~\citep{MCCLOSKEY1989109, kirkpatrick2017overcoming} is a central challenge when adapting pretrained models to new languages. \citet{khelli-etal-2025-causes} find that partial parameter sharing can mitigate forgetting in multilingual CPT, while \citet{owodunni2025continuallyaddingnewlanguages} study layer-selective fine-tuning but find no clear advantage of parameter-efficient methods over full fine-tuning. However, these methods typically constrain parameters uniformly, or in regions chosen apriori. Our work instead looks into which layers should be preserved and whether the answer depends on the downstream task.

\textbf{Layer Specialization in Multilingual Models}\quad
~\citet{wendler-etal-2024-llamas, dumas-etal-2025-separating} suggest that intermediate representations of a model are biased toward English, with language-specific processing concentrated in early and middle layers~\citep{tang2024languagespecificneuronskeymultilingual,kojima2024multilingualabilitydecoderbasedpretrained,NEURIPS2024_1bd359b3}. Another approach that combines language and mathematical-reasoning experts through layer swapping~\citep{bandarkar2025layerswappingzeroshotcrosslingual}. \citet{bandarkar-peng-2025-unreasonable} also find that fine-tuning all parameters and then reverting selected updates can outperform freezing them during fine-tuning. These studies focus mostly on transferring reasoning through fine-tuning, with limited exploration of CPT. We extend this line of work to multilingual CPT and find that this phenomenon is task-dependent, as the middle-layer region whose reversion best recovers reading comprehension harms translation, and post-hoc reversion benefits do not reliably carry over to other preservation methods during training.

\textbf{Expert Language Modeling}\quad Grouping languages by family can reduce cross-lingual interference and improve transfer to related low-resource languages~\citep{chronopoulou-etal-2023-language,ogueji-etal-2021-small,downey-etal-2024-targeted}. \textsc{x-ELM}~\citep{blevins-etal-2024-breaking} extends Branch-Train-Merge~\citep{li2022branchtrainmergeembarrassinglyparalleltraining} to multilingual pretraining by training experts on typologically clustered languages subsets that outperform compute-matched dense model. We adopt family experts as one of our CPT settings and find that specialization is task-dependent which is experts forget less than dense CPT on comprehension but under-perform on translation.

\section{Conclusion}

We investigate how layer-level parameter updates shape the acquisition--forgetting knowledge tradeoff during multilingual CPT. Specifically, we perform a layer-group interpolation analysis to uncover which layers most affect task performance change after jointly finetuning a model on five language families, resulting in decreased performance for reading comprehension but much stronger translation performance. Our analysis shows that the layers causing these performance changes are task-dependent: reverting the middle layers yields the largest gains in comprehension performance, while translation generally exhibits \textit{worse} performance when reverting to the base model, though exact trends vary across language families and translation direction. Following these findings, targeted freezing and regularization interventions during CPT at these layers improve comprehension performance when training language family expert models over that base, but these gains do not consistently transfer to translation (where jointly fine-tuning with layer reversion performs the best). 

These findings caution us against generalizing the insights from previous work with respect to layer-wise effects on reasoning tasks to other tasks. Instead, we observe that a layer range whose reversion benefits one task (comprehension) can impair another (translation), with the effect of a given layer range likely depending on underlying task features such as whether it is discriminative or generative, monolingual or cross-lingual, and the type of knowledge required. Thus, we propose our interpolation-based localization approach as a diagnostic framework to determine which layers (if any) should be preserved during task- and language-specific CPT, rather than positioning our results as a source of universal, task-agnostic guidelines.
\section*{Acknowledgments}
We would like to thank Eugene Jang for feedback on the initial project idea and giving detailed and helpful comments on our draft. We would also like to thank Sanjana Londhe who helped us in designing the Figure \ref{fig:method} for our draft. 

This work used H200 GPUs at NCSA DeltaAI through allocation CIS251341 from the Advanced Cyberinfrastructure Coordination Ecosystem: Services \& Support (ACCESS) program, which is supported by U.S. National Science Foundation grants \#2138259, \#2138286, \#2138307, \#2137603, and \#2138296 \citep{10.1145/3569951.3597559}.

\subsection*{AI use statement}
In this work, we used generative AI for assistance with writing code. We used AI-assisted literature review tools to supplement our manual literature reviewing process. We also used coding agents for assistance in running our experiments, with human in loop. AI was used for typesetting and generating LaTeX tables for our paper and minor writing refinements, with manual verification. We take responsibility for the final content of this work, including text, claims or artifacts produced with the aid of generative AI.

\raggedright
\bibliography{iclr2027_conference}
\bibliographystyle{iclr2027_conference}
\clearpage
\appendix
\onecolumn
\section{Appendix}

\subsection{Limitations}
\label{sec:limitations}

All experiments use a single 4B-parameter model (Gemma-3 4B) with a fixed budget of 5B tokens per family from one web corpus (MADLAD-400); we do not evaluate whether the localization results or intervention effects transfer to other model scales, architectures, or data regimes.
The preservation techniques we evaluate (freezing, L2-SP, post-hoc reversion, and souping) are established methods; our contribution is using interpolation-based localization to select where to apply them, and testing whether the selected layers transfer across tasks and from post-hoc to training-time interventions.
The localization analysis is performed only on the Dense CPT checkpoint, and the FLORES window is selected on a 20-direction screen of that checkpoint before being evaluated on the full held-in FLORES sets.

\subsection{Held-Out Evaluation Languages}

\label{app:held_out_langs}

Table~\ref{tab:held_out_languages} lists all held-out languages evaluated in \S\ref{sec:heldout}, grouped by family, together with their benchmark coverage.
These languages are excluded from the CPT training set but are related to the training languages of the corresponding family group, allowing us to probe within-family generalization. The Slavic group also includes Lithuanian and Latvian, which are Baltic languages closely related to, but not members of, the Slavic family.

\subsection{Training Hyperparameters and Fairness Controls}
\label{app:training_details}

Tables~\ref{tab:dense_cpt_hparams} and~\ref{tab:expert_hparams} summarize the optimization settings used for the Gemma-3 4B experiments. Post-hoc reversion (D.-Rev.\ and E.-Rev.) and Expert Soup reuse the trained Dense and vanilla Expert checkpoints and introduce no additional optimization steps.

\textbf{Targeted window constraints.}\quad For a selected six-layer window $G$ ($[10,16)$ or $[5,11)$), Layer Freezing holds all parameters in $G$ fixed at their Base values and trains every other layer. Layer-Range L2-SP trains all parameters and applies the penalty $\lambda=0.05$ only to parameters in $G$, with $\lambda=0$ elsewhere. We keep the learning rate unchanged across all family-specific variants.

\textbf{Layer-Range L2-SP $\lambda$ selection.}\quad We use a single penalty strength, $\lambda=0.05$, for both windows and all families, without per-window or per-family tuning. Other values may change the balance between preservation and acquisition, particularly for translation.
\begin{table*}[t]
\centering
\small
\begin{tabular}{ll l ccc}
\toprule
\textbf{Family} & \textbf{Code} & \textbf{Language} & \textbf{PPL} & \textbf{Belebele} & \textbf{FLORES} \\
\midrule
\multirow{6}{*}{Slavic}
  & bg & Bulgarian  & \ding{51} & \ding{51} & \ding{51} \\
  & cs & Czech      & \ding{51} & \ding{51} & \ding{51} \\
  & lt & Lithuanian & \ding{51} & \ding{51} & \ding{51} \\
  & pl & Polish     & \ding{51} & \ding{51} & \ding{51} \\
  & sl & Slovenian  & \ding{51} & \ding{51} & \ding{51} \\
  & lv & Latvian    & \ding{51} & \ding{51} & \ding{51} \\
\midrule
\multirow{3}{*}{Germanic}
  & is & Icelandic  & \ding{51} & \ding{51} & \ding{51} \\
  & no & Norwegian  & \ding{51} & \ding{51} & \ding{51} \\
  & sv & Swedish    & \ding{51} & \ding{51} & \ding{51} \\
\midrule
\multirow{7}{*}{Indic}
  & as & Assamese   & \ding{51} & \ding{51} & \ding{51} \\
  & gu & Gujarati   & \ding{51} & \ding{51} & \ding{51} \\
  & or & Odia       & \ding{51} & \ding{51} & \ding{51} \\
  & pa & Punjabi    & \ding{51} & \ding{51} & \ding{51} \\
  & sd & Sindhi     & \ding{51} & \ding{51} & \ding{51} \\
  & si & Sinhala    & \ding{51} & \ding{51} & \ding{51} \\
  & ur & Urdu       & \ding{51} & \ding{51} & \ding{51} \\
\midrule
\multirow{5}{*}{Austronesian}
  & ilo & Ilocano   & \ding{51} & \ding{51} & \ding{51} \\
  & mi  & M\={a}ori & \ding{51} & \ding{51} & \ding{51} \\
  & su  & Sundanese & \ding{51} & \ding{51} & \ding{51} \\
  & war & Waray     & \ding{51} & \ding{51} & \ding{51} \\
  & mg  & Malagasy  & \ding{51} & \ding{51} & \ding{51} \\
\midrule
Romance
  & ca & Catalan    & \ding{51} & \ding{51} & \ding{51} \\
\bottomrule
\end{tabular}
\caption{Held-out evaluation languages per benchmark. \ding{51}~= evaluated. PPL = held-out perplexity; FLORES results cover both \textsc{en}$\to$\textsc{xx} and \textsc{xx}$\to$\textsc{en} directions.}
\label{tab:held_out_languages}
\end{table*}

\begin{table}[h]
\centering
\small
\begin{tabular}{@{}p{4.2cm} p{8.5cm}@{}}
\toprule
\textbf{Hyperparameter} & \textbf{Dense CPT} \\
\midrule
Training data & All five families \\
Corpus budget & 25B tokens total \\
Maximum steps & 50{,}000 \\
Learning rate & $5{\times}10^{-5}$ \\
Warmup / scheduler & 500 steps / cosine \\
Batching & Per-device batch size 4, gradient accumulation 12, sequence length 2048, packed sequences \\
Validation & Every 1000 steps \\
Early stopping & Patience 3, threshold 0.001 \\
Optimizer defaults & Weight decay 0.01, max gradient norm 1.0, seed 42, bfloat16, gradient checkpointing \\
Layer treatment & All layers trainable \\
\bottomrule
\end{tabular}
\caption{Training configuration for Dense CPT on Gemma-3 4B.}
\label{tab:dense_cpt_hparams}
\end{table}

\begin{table}[h]
\centering
\small
\begin{tabular}{@{}p{3.8cm} p{3cm} p{3cm} p{4.5cm}@{}}
\toprule
\textbf{Hyperparameter} & \textbf{Family Expert} & \textbf{Layer Freezing} & \textbf{Layer-Range L2-SP} \\
\midrule
Training data & One family & One family & One family \\
Corpus budget & 5B tokens/family & 5B tokens/family & 5B tokens/family \\
Maximum steps & 16{,}954 & \multicolumn{2}{p{7.5cm}@{}}{18{,}000 or 21{,}000 (fixed budget of ${\approx}0.35$ epoch)$^\dagger$} \\
Learning rate & $2{\times}10^{-5}$ & $2{\times}10^{-5}$ & $2{\times}10^{-5}$ \\
Warmup / scheduler & 200 steps / cosine & 200 steps / cosine & 200 steps / cosine \\
Batching & \multicolumn{3}{p{10.5cm}@{}}{Per-device batch size 4, gradient accumulation 12, sequence length 2048, packed sequences} \\
Validation & Every 500 steps & Every 500 steps & Every 500 steps \\
Early stopping & Patience 6, threshold 0.005 & \multicolumn{2}{p{7.5cm}@{}}{Patience 6, threshold 0.005} \\
Optimizer defaults & \multicolumn{3}{p{10.5cm}@{}}{Weight decay 0.01, max gradient norm 1.0, seed 42, bfloat16, gradient checkpointing} \\
Layer treatment & All layers trainable & Freeze the six layers in $G$; train all other layers & L2-SP on layers in $G$ only; $\lambda{=}0.05$ in $G$, $\lambda{=}0$ elsewhere \\
\bottomrule
\end{tabular}
\caption{Training configuration for family-specific expert variants on Gemma-3 4B. $G$ is the targeted window, $[10,16)$ or $[5,11)$. $^\dagger$21{,}000 steps for Romance (both windows) and for Indic and Austronesian $[10,16)$; 18{,}000 steps otherwise.}
\label{tab:expert_hparams}
\end{table}

\subsection{Granular Sliding Window Results}
\label{app:causal_layer_sweep}

Tables~\ref{tab:layer_window_belebele_family} and~\ref{tab:layer_window_flores_family_direction} provide the corresponding family-wise breakdown for the fixed-width window reversions in Table~\ref{tab:layer_window_belebele}.

\begin{table}[t]
\centering
\scriptsize
\begin{tabular}{@{}lrrrrr@{}}
\toprule
\textbf{Layers reverted} & \textbf{Slavic} & \textbf{Germanic} & \textbf{Indic} & \textbf{Austronesian} & \textbf{Romance} \\
\midrule
$[0,6)$   & $+1.5$ & $+3.0$ & $+1.6$ & $+1.5$ & $+3.1$ \\
$[5,11)$  & $+3.4$ & $+4.1$ & $+3.1$ & $+5.0$ & $+4.7$ \\
$[10,16)$ & \textbf{$+5.6$} & \textbf{$+6.5$} & \textbf{$+4.9$} & \textbf{$+5.5$} & \textbf{$+7.1$} \\
$[24,30)$ & $-0.7$ & $-0.5$ & $-0.5$ & $+0.4$ & $-0.5$ \\
$[27,33)$ & $+0.4$ & $+0.2$ & $+0.1$ & $+0.9$ & $+0.5$ \\
$[28,34)$ & $+0.3$ & $+0.8$ & $+0.5$ & $+1.0$ & $+1.0$ \\
\bottomrule
\end{tabular}
\caption{Family-wise Belebele accuracy changes (percentage points) relative to Dense CPT for the selected fixed-width layer reversions. Each family score averages its held-in evaluation languages.}
\label{tab:layer_window_belebele_family}
\end{table}

\begin{table}[t]
\centering
\small
\begin{tabular}{@{}llrr@{}}
\toprule
\textbf{Layers reverted} & \textbf{Family} & \textbf{xx$\to$en $\Delta$ ChrF} & \textbf{en$\to$xx $\Delta$ ChrF} \\
\midrule
\multirow{5}{*}{$[0,6)$}
 & Slavic & $+3.14$ & $+2.13$ \\
 & Germanic & $-1.55$ & $+0.75$ \\
 & Indic & $-0.19$ & $-0.65$ \\
 & Austronesian & $-4.14$ & $-0.48$ \\
 & Romance & $+3.38$ & $+0.18$ \\
\midrule
\multirow{5}{*}{$[5,11)$}
 & Slavic & $+3.08$ & $+1.92$ \\
 & Germanic & $+1.61$ & $+0.68$ \\
 & Indic & $-5.24$ & $-0.75$ \\
 & Austronesian & $+1.01$ & $+0.32$ \\
 & Romance & $+3.86$ & $+1.01$ \\
\midrule
\multirow{5}{*}{$[10,16)$}
 & Slavic & $-11.03$ & $-0.51$ \\
 & Germanic & $-7.53$ & $-0.10$ \\
 & Indic & $-2.89$ & $-1.27$ \\
 & Austronesian & $-3.97$ & $-0.04$ \\
 & Romance & $-6.62$ & $-0.76$ \\
\midrule
\multirow{5}{*}{$[24,30)$}
 & Slavic & $-8.50$ & $-0.23$ \\
 & Germanic & $-4.89$ & $-0.76$ \\
 & Indic & $-0.03$ & $-1.38$ \\
 & Austronesian & $-3.41$ & $-17.03$ \\
 & Romance & $-5.82$ & $-0.70$ \\
\midrule
\multirow{5}{*}{$[27,33)$}
 & Slavic & $-24.63$ & $-10.35$ \\
 & Germanic & $-29.91$ & $-9.42$ \\
 & Indic & $-12.19$ & $-12.15$ \\
 & Austronesian & $-23.52$ & $-28.82$ \\
 & Romance & $-27.38$ & $-6.68$ \\
\midrule
\multirow{5}{*}{$[28,34)$}
 & Slavic & $-11.98$ & $-4.95$ \\
 & Germanic & $-8.11$ & $-7.01$ \\
 & Indic & $+1.79$ & $-7.44$ \\
 & Austronesian & $-7.95$ & $-20.63$ \\
 & Romance & $-11.14$ & $-2.57$ \\
\bottomrule
\end{tabular}
\caption{Family- and direction-wise FLORES ChrF changes relative to Dense CPT for the selected six-layer Dense-to-Base reversions. Each cell averages two held-in languages from the 20-direction screen. The Dense reference is the mean of the three archived Dense evaluations, rescored with the first-line post-truncation protocol.}
\label{tab:layer_window_flores_family_direction}
\end{table}

\begin{figure}[h]
\centering
\includegraphics[width=0.7\textwidth]{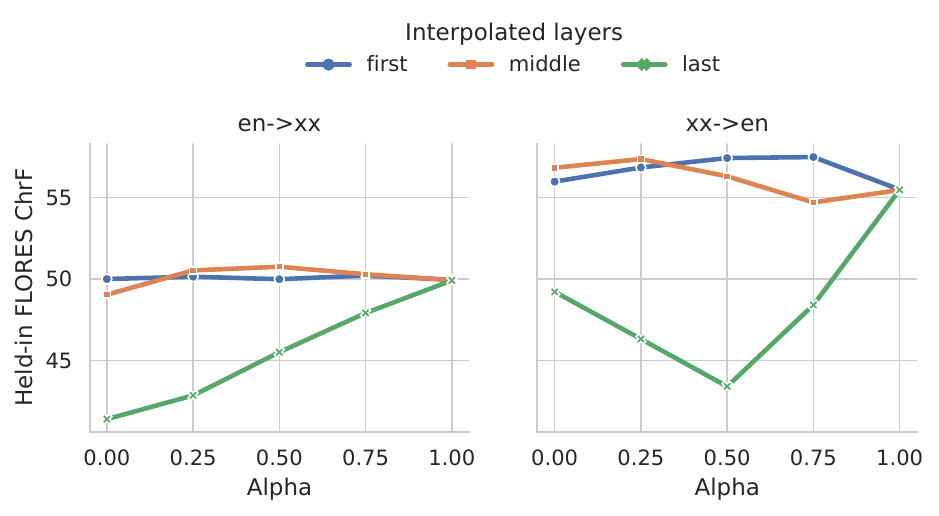}
\caption{Held-in FLORES ChrF under layer interpolation between the base model and Dense CPT, reported separately for en$\to$xx and xx$\to$en directions. All non-interpolated layers are kept at their Dense CPT values.}
\label{fig:flores_layer_sweep_overall}
\end{figure}

\begin{figure*}[h]
\centering
\includegraphics[width=\textwidth]{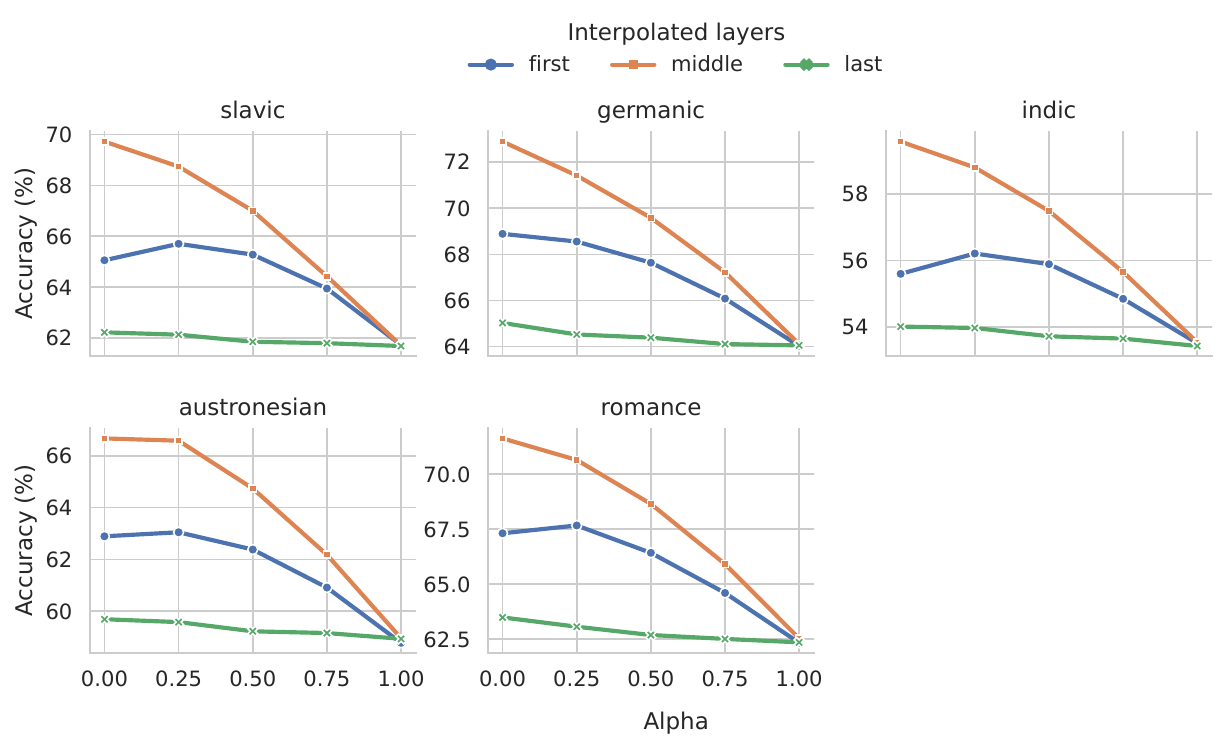}
\caption{Held-in Belebele accuracy under first-, middle-, and last-layer interpolation, broken down by language family.}
\label{fig:causal_layer_belebele_sweep_family}
\end{figure*}

\subsection{Held-Out Family-Level Results}
\label{app:heldout_ppl}

Table~\ref{tab:perplexity_heldout} reports targeted-window perplexity on held-out relatives.

Table~\ref{tab:belebele_targeted_heldout} gives the corresponding held-out Belebele family averages for the targeted $[10,16)$ interventions.

\begin{table*}[t]
\centering
\scriptsize
\begin{tabular}{l rrrr rrrr rrrr}
\toprule
& \multicolumn{4}{c}{\textbf{Shared baselines}} & \multicolumn{4}{c}{$\mathbf{[5,11)}$} & \multicolumn{4}{c}{$\mathbf{[10,16)}$} \\
\cmidrule(lr){2-5} \cmidrule(lr){6-9} \cmidrule(lr){10-13}
\textbf{Family} & \textbf{Base} & \textbf{Dense} & \textbf{Expert} & \textbf{Soup} & \textbf{D.-Rev.} & \textbf{E.-Rev.} & \textbf{Freeze} & \textbf{L2-SP} & \textbf{D.-Rev.} & \textbf{E.-Rev.} & \textbf{Freeze} & \textbf{L2-SP} \\
\midrule
Slavic & 7.36 & 9.16 & 7.64 & \textbf{7.10} & 9.26 & 7.67 & 7.63 & 7.65 & 9.12 & 7.59 & 7.63 & 7.68 \\
Germanic & 8.56 & 11.44 & 9.23 & \textbf{8.30} & 11.26 & 9.23 & 9.17 & 9.18 & 11.33 & 9.09 & 9.18 & 9.15 \\
Indic & 8.99 & 14.70 & 9.76 & \textbf{8.94} & 14.65 & 9.79 & 9.78 & 9.79 & 14.26 & 9.65 & 9.86 & 9.76 \\
Austronesian & \textbf{14.04} & 28.36 & 28.35 & 14.17 & 29.93 & 30.00 & 29.99 & 30.25 & 28.75 & 25.08 & 27.69 & 27.58 \\
Romance & 7.90 & 9.07 & 7.93 & \textbf{7.57} & 9.31 & 8.01 & 7.93 & 7.89 & 9.21 & 7.94 & 7.97 & 7.97 \\
\bottomrule
\end{tabular}
\caption{Held-out validation perplexity $\downarrow$, averaged over held-out relatives within each family. D.-Rev. restores Dense layers in the indicated window to Base; Freeze and L2-SP preserve the indicated layers during family CPT; E.-Rev. restores them to Base after Expert CPT. Germanic excludes German. \textbf{Bold} = lowest value within a row.}
\label{tab:perplexity_heldout}
\end{table*}

\begin{table*}[t]
\centering
\scriptsize
\begin{tabular}{l rrrr rrrr}
\toprule
& \multicolumn{4}{c}{\textbf{Shared baselines}} & \multicolumn{4}{c}{$\mathbf{[10,16)}$} \\
\cmidrule(lr){2-5} \cmidrule(lr){6-9}
\textbf{Family} & \textbf{Base} & \textbf{Dense} & \textbf{Expert} & \textbf{Soup} & \textbf{D.-Rev.} & \textbf{E.-Rev.} & \textbf{Freeze} & \textbf{L2-SP} \\
\midrule
Slavic & \textbf{0.725} & 0.598 & 0.710 & 0.719 & 0.657 & 0.721 & 0.708 & 0.716 \\
Germanic & \textbf{0.722} & 0.592 & 0.699 & 0.716 & 0.652 & 0.708 & 0.713 & 0.709 \\
Indic & \textbf{0.576} & 0.476 & 0.557 & 0.573 & 0.521 & 0.566 & 0.564 & 0.566 \\
Austronesian & \textbf{0.487} & 0.383 & 0.432 & 0.475 & 0.399 & 0.440 & 0.456 & 0.455 \\
Romance & \textbf{0.752} & 0.600 & 0.731 & \textbf{0.752} & 0.682 & 0.744 & 0.741 & 0.744 \\
\midrule
Average & \textbf{0.652} & 0.530 & 0.626 & 0.647 & 0.582 & 0.636 & 0.636 & 0.638 \\
\bottomrule
\end{tabular}
\caption{Held-out Belebele accuracy $\uparrow$ (2-shot), averaged over held-out relatives within each family; Average weights families equally. D.-Rev. restores Dense layers $[10,16)$ to Base; Freeze and L2-SP preserve $[10,16)$ during family CPT; E.-Rev. restores $[10,16)$ to Base after Expert CPT. Germanic excludes German. \textbf{Bold} = best value within a row.}
\label{tab:belebele_targeted_heldout}
\end{table*}

\clearpage

\subsection{FLORES-200 Evaluation Protocol}
\label{app:flores_eval_protocol}

We evaluate FLORES-200 with lm-eval-harness in a 2-shot setting and report corpus ChrF for both \textsc{en}$\to$\textsc{xx} and \textsc{xx}$\to$\textsc{en} directions. Our initial goal was to match the Gemma-3 technical report's FLORES setup as closely as possible for the shared base model~\citep{gemmateam2025gemma3technicalreport}. However, exact replication was not possible with the default lm-eval-harness task configuration, since newline stopping could terminate some base-model generations before a translation was produced.

We therefore use a uniform post-truncation protocol for all checkpoints. After generation, we strip leading whitespace, keep only the text before the first generated newline or literal \texttt{\textbackslash n}, and score the remaining span against the reference with ChrF. This keeps decoding and scoring comparable across the base model, Dense CPT, family experts, reverted checkpoints, Layer Freezing, Layer-Range L2-SP, and Expert Soup. The resulting FLORES numbers should therefore be read as a controlled comparison under a shared evaluation pipeline, rather than as a direct reproduction of the Gemma-3 technical report score.


\subsection{Targeted FLORES Results and Comparison Details}
\label{app:flores_targeted}

Table~\ref{tab:flores_targeted_direction} separates the $[5,11)$ results in Table~\ref{tab:flores_targeted_training} by translation direction.

\textbf{Empty outputs.}\quad Empty-output rates provide an important diagnostic alongside ChrF. All Austronesian family-expert variants have high rates: $57.5\%$ for the vanilla Expert, $62.55\%$ for E.-Rev., $58.97\%$ for Freeze, and $61.37\%$ for L2-SP, compared with at most $3.69\%$ for any other family-specific checkpoint. Their translation scores therefore reflect substantial output failures as well as the quality of nonempty translations. The final block of Table~\ref{tab:flores_targeted_direction} shows that reverting the Austronesian expert's $[27,33)$ layers to Base reduces the empty-output rate to $7.0\%$ (\S\ref{sec:flores}), from $57.5\%$ for the archived Expert evaluation and $38.9\%$ for the separate Expert generation run used as that block's reference; here $\beta$ is the fraction of the way from Expert to Base weights, so $\beta=1$ corresponds to $\alpha=0$ in the notation of \S\ref{sec:causal_analysis}.

\begin{table}[t]
\centering
\small
\begin{tabular}{@{}llrrr@{}}
\toprule
\textbf{Group} & \textbf{Method} & \textbf{\textsc{xx}$\to$\textsc{en}} & \textbf{\textsc{en}$\to$\textsc{xx}} & \textbf{Empty (\%)} \\
\midrule
Slavic & Dense & 56.74 & 48.66 & 1.38 \\
 & D.-Rev. & 59.80 & 49.64 & 0.02 \\
 & E.-Rev. & 52.98 & 49.66 & 0.17 \\
 & Freeze & 45.99 & 48.24 & 0.19 \\
 & L2-SP & 45.34 & 48.38 & 0.33 \\
\midrule
Germanic & Dense & 61.39 & 56.64 & 0.56 \\
 & D.-Rev. & 64.73 & 57.80 & 0.00 \\
 & E.-Rev. & 61.48 & 57.82 & 1.21 \\
 & Freeze & 58.53 & 56.91 & 1.31 \\
 & L2-SP & 58.91 & 57.23 & 1.82 \\
\midrule
Indic & Dense & 47.68 & 39.16 & 0.15 \\
 & D.-Rev. & 42.79 & 39.75 & 0.02 \\
 & E.-Rev. & 34.28 & 38.48 & 3.16 \\
 & Freeze & 34.61 & 38.09 & 2.16 \\
 & L2-SP & 35.56 & 37.61 & 1.86 \\
\midrule
Austronesian & Dense & 57.22 & 53.87 & 0.95 \\
 & D.-Rev. & 59.58 & 52.90 & 0.08 \\
 & E.-Rev. & 14.24 & 30.11 & 62.55 \\
 & Freeze & 17.88 & 31.46 & 58.97 \\
 & L2-SP & 17.50 & 29.28 & 61.37 \\
\midrule
Romance & Dense & 58.61 & 57.12 & 0.68 \\
 & D.-Rev. & 62.72 & 57.75 & 0.04 \\
 & E.-Rev. & 55.05 & 57.71 & 3.69 \\
 & Freeze & 47.48 & 51.42 & 3.47 \\
 & L2-SP & 46.91 & 50.13 & 3.18 \\
\midrule
\multicolumn{5}{l}{\textit{Austronesian: Expert-to-Base interpolation in $[27,33)$}} \\
 & Expert (re-run)$^\ast$ & 29.05 & 30.36 & 38.9 \\
 & $\beta=0.5$ & 31.40 & 39.17 & 35.3 \\
 & $\beta=1.0$ & 40.02 & 40.61 & 7.0 \\
\bottomrule
\end{tabular}
\caption{Direction-wise held-in FLORES ChrF for the pipeline-matched Dense/D.-Rev.\ pair and the $[5,11)$ family-specific interventions in Table~\ref{tab:flores_targeted_training}. Dense is the pipeline-matched control ($\alpha=1$) for D.-Rev. Scores use first-line post-truncation scoring and are macro-averaged within each group. Empty-output percentages aggregate both directions. The additional Austronesian block interpolates the vanilla expert's $[27,33)$ layers toward Base, with $\beta=1$ denoting full replacement (equivalent to $\alpha=0$). These interventions are separate from the Freeze and L2-SP checkpoints above. $^\ast$A separate generation run of the vanilla Austronesian expert, used as the paired reference for this block; the Expert evaluation in Table~\ref{tab:flores_targeted_training} scores $19.28$ / $30.75$ ChrF with a $57.5\%$ empty-output rate.}
\label{tab:flores_targeted_direction}
\end{table}








\subsection{Per-Language Results Tables}
\label{app:perlang}

Tables~\ref{tab:perplexity_perlang}--\ref{tab:flores_en_xx_perlang} report per-language results for perplexity, Belebele, and FLORES.
For each language, the Expert, E.-Rev., Freeze, and L2-SP columns report the model trained on that language's family (e.g., the Slavic expert for Croatian and Russian).
All other strategy columns (Dense, D.-Rev., Soup) are single models evaluated across all languages.

\begin{table}[p]
\centering
\scriptsize
\resizebox{\textwidth}{!}{%
\begin{tabular}{llrrrrrrrr}
\toprule
\textbf{Family} & \textbf{Language} & \textbf{Base} & \textbf{Dense} & \textbf{Expert} & \textbf{Soup} & \textbf{D.-Rev.$_{10}$} & \textbf{E.-Rev.$_{10}$} & \textbf{Freeze$_{10}$} & \textbf{L2-SP$_{10}$} \\
\midrule
Slavic & Macedonian (mk) & 6.70 & \textbf{4.96} & \underline{5.06} & 6.11 & \underline{5.10} & 5.20 & \underline{5.15} & 5.20 \\
Slavic & Croatian (hr) & 8.42 & \textbf{6.90} & \textbf{6.90} & 7.90 & \underline{7.09} & 7.11 & \underline{7.01} & \underline{7.09} \\
Slavic & Russian (ru) & 10.37 & \underline{9.41} & \textbf{9.21} & 9.91 & 9.58 & \underline{9.41} & \underline{9.29} & \underline{9.38} \\
Slavic & Slovak (sk) & 7.25 & \underline{6.14} & \textbf{6.12} & 6.72 & \underline{6.30} & \underline{6.29} & \underline{6.19} & \underline{6.26} \\
Slavic & Serbian (sr) & 7.03 & \textbf{5.45} & \underline{5.58} & 6.57 & \underline{5.62} & 5.73 & 5.69 & 5.73 \\
Slavic & Ukrainian (uk) & 6.20 & \underline{5.59} & \textbf{5.51} & 5.89 & \underline{5.71} & \underline{5.65} & \underline{5.55} & \underline{5.61} \\
\midrule
Germanic & Afrikaans (af) & 11.24 & \textbf{7.41} & 7.71 & 9.92 & 7.72 & 7.98 & 8.03 & 8.04 \\
Germanic & Frisian (fy) & 14.89 & \textbf{4.86} & 5.28 & 11.20 & 5.35 & 5.76 & 5.69 & 5.70 \\
Germanic & Luxembourgish (lb) & 14.32 & \textbf{5.61} & 6.08 & 11.77 & 6.10 & 6.50 & 6.48 & 6.49 \\
Germanic & Danish (da) & 7.42 & \underline{6.14} & \textbf{6.09} & 6.92 & 6.36 & 6.30 & \underline{6.22} & \underline{6.22} \\
Germanic & Dutch (nl) & 8.69 & \underline{7.53} & \textbf{7.40} & 8.18 & 7.77 & 7.66 & \underline{7.55} & \underline{7.55} \\
Germanic & English (en) & 11.91 & 11.51 & \textbf{10.83} & 11.42 & 11.63 & \underline{10.99} & \underline{10.94} & \underline{10.94} \\
\midrule
Indic & Bengali (bn) & 13.47 & \underline{11.05} & \textbf{10.99} & 12.31 & 11.49 & 11.40 & 11.34 & 11.29 \\
Indic & Hindi (hi) & 10.17 & \underline{8.47} & \textbf{8.39} & 9.52 & 8.81 & 8.70 & 8.62 & \underline{8.59} \\
Indic & Kannada (kn) & 7.73 & \textbf{5.35} & \underline{5.49} & 6.53 & 5.58 & 5.72 & 5.71 & 5.71 \\
Indic & Malayalam (ml) & 7.94 & \textbf{5.69} & \underline{5.79} & 6.73 & 5.94 & 6.05 & 6.03 & 6.01 \\
Indic & Marathi (mr) & 10.83 & \textbf{8.25} & \underline{8.28} & 9.44 & 8.62 & 8.65 & 8.57 & 8.55 \\
Indic & Nepali (ne) & 11.72 & \textbf{8.11} & 8.40 & 9.95 & 8.42 & 8.68 & 8.74 & 8.73 \\
Indic & Tamil (ta) & 11.49 & \textbf{8.78} & \underline{8.88} & 9.91 & 9.23 & 9.30 & 9.24 & 9.21 \\
Indic & Telugu (te) & 7.16 & \textbf{5.54} & \underline{5.58} & 6.20 & 5.76 & 5.81 & 5.77 & 5.76 \\
\midrule
Austronesian & Samoan (sm) & 11.39 & \textbf{4.00} & 4.65 & 9.39 & 4.61 & 5.08 & 5.09 & 5.10 \\
Austronesian & Javanese (jv) & 17.58 & \textbf{6.45} & 7.69 & 14.36 & 7.64 & 8.49 & 8.62 & 8.64 \\
Austronesian & Cebuano (ceb) & 10.12 & \textbf{4.63} & 5.29 & 8.72 & 5.28 & 5.75 & 5.82 & 5.82 \\
Austronesian & Filipino (fil) & 10.10 & \textbf{5.32} & 5.61 & 8.54 & 6.18 & 6.41 & 5.94 & 5.93 \\
Austronesian & Indonesian (id) & 10.89 & \textbf{8.74} & \underline{8.85} & 9.99 & 9.24 & 9.27 & 9.15 & 9.13 \\
Austronesian & Malay (ms) & 10.60 & \textbf{8.20} & \underline{8.38} & 9.59 & 8.49 & 8.63 & 8.60 & 8.59 \\
\midrule
Romance & Spanish (es) & 10.81 & 9.93 & \textbf{9.63} & 10.19 & 10.13 & 9.84 & \underline{9.83} & 9.90 \\
Romance & Portuguese (pt) & 10.12 & 9.09 & \textbf{8.81} & 9.48 & 9.27 & 9.04 & \underline{9.01} & 9.06 \\
Romance & French (fr) & 9.47 & 8.55 & \textbf{8.33} & 8.78 & 8.79 & 8.56 & 8.56 & 8.61 \\
Romance & Galician (gl) & 11.19 & \textbf{8.21} & \underline{8.37} & 10.06 & 8.44 & 8.55 & 8.60 & 8.66 \\
Romance & Italian (it) & 10.77 & \underline{8.94} & \textbf{8.77} & 10.03 & 9.54 & 9.31 & 9.09 & 9.14 \\
Romance & Romanian (ro) & 7.67 & \underline{6.56} & \textbf{6.50} & 7.16 & 6.76 & \underline{6.69} & \underline{6.69} & 6.72 \\
\bottomrule
\multicolumn{10}{l}{}
\end{tabular}}
\caption{Per-language perplexity $\downarrow$ on held-in validation text. D.-Rev.$_{10}$ and E.-Rev.$_{10}$ restore layers $[10,16)$ of the Dense and Expert models to Base; Freeze$_{10}$ and L2-SP$_{10}$ use the same targeted layer window. Expert, E.-Rev.$_{10}$, and the targeted training columns report the model trained on the language's family. \textbf{Bold} = best per row; \underline{underline} = within 0.2 of best.}
\label{tab:perplexity_perlang}
\end{table}

\begin{table}[p]
\centering
\scriptsize
\resizebox{\textwidth}{!}{%
\begin{tabular}{llrrrrrrrr}
\toprule
\textbf{Family} & \textbf{Language} & \textbf{Base} & \textbf{Dense} & \textbf{Expert} & \textbf{Soup} & \textbf{D.-Rev.$_{10}$} & \textbf{E.-Rev.$_{10}$} & \textbf{Freeze$_{10}$} & \textbf{L2-SP$_{10}$} \\
\midrule
Slavic & Macedonian (mk) & 0.684 & 0.626 & 0.702 & 0.696 & 0.654 & \textbf{0.719} & 0.709 & 0.717 \\
Slavic & Croatian (hr) & 0.727 & 0.606 & 0.736 & \textbf{0.751} & 0.666 & 0.736 & 0.737 & 0.738 \\
Slavic & Russian (ru) & 0.730 & 0.606 & 0.720 & \textbf{0.746} & 0.664 & 0.737 & 0.722 & 0.727 \\
Slavic & Slovak (sk) & 0.744 & 0.638 & 0.737 & \textbf{0.751} & 0.681 & 0.744 & \textbf{0.751} & 0.749 \\
Slavic & Serbian (sr) & 0.708 & 0.628 & 0.736 & 0.739 & 0.699 & \textbf{0.743} & 0.729 & 0.731 \\
Slavic & Ukrainian (uk) & 0.722 & 0.610 & 0.726 & 0.724 & 0.687 & \textbf{0.728} & 0.714 & 0.722 \\
\midrule
Germanic & Afrikaans (af) & 0.734 & 0.617 & \textbf{0.747} & 0.742 & 0.684 & 0.743 & 0.739 & \textbf{0.747} \\
Germanic & Danish (da) & 0.743 & 0.624 & 0.740 & \textbf{0.756} & 0.699 & 0.748 & 0.750 & \textbf{0.756} \\
Germanic & Dutch (nl) & 0.742 & 0.653 & 0.742 & \textbf{0.748} & 0.692 & 0.741 & 0.739 & 0.744 \\
Germanic & English (en) & 0.813 & 0.674 & 0.793 & 0.798 & 0.754 & 0.803 & \textbf{0.820} & 0.807 \\
\midrule
Indic & Bengali (bn) & 0.654 & 0.550 & 0.616 & 0.653 & 0.603 & 0.658 & 0.657 & \textbf{0.659} \\
Indic & Hindi (hi) & 0.514 & 0.539 & 0.567 & \textbf{0.616} & 0.579 & 0.612 & 0.603 & 0.608 \\
Indic & Kannada (kn) & 0.601 & 0.518 & 0.569 & 0.602 & 0.563 & 0.592 & 0.600 & \textbf{0.610} \\
Indic & Malayalam (ml) & 0.632 & 0.532 & 0.616 & \textbf{0.664} & 0.603 & 0.654 & 0.659 & 0.651 \\
Indic & Marathi (mr) & 0.649 & 0.539 & 0.617 & 0.654 & 0.596 & 0.644 & 0.650 & \textbf{0.661} \\
Indic & Nepali (ne) & 0.574 & 0.547 & 0.564 & \textbf{0.624} & 0.584 & 0.621 & 0.602 & 0.616 \\
Indic & Tamil (ta) & 0.617 & 0.543 & 0.613 & \textbf{0.628} & 0.581 & 0.623 & 0.623 & 0.614 \\
Indic & Telugu (te) & 0.568 & 0.509 & 0.584 & 0.592 & 0.564 & 0.584 & 0.587 & \textbf{0.597} \\
\midrule
Austronesian & Javanese (jv) & 0.542 & 0.522 & 0.569 & 0.594 & 0.578 & 0.606 & \textbf{0.637} & 0.631 \\
Austronesian & Cebuano (ceb) & 0.658 & 0.570 & 0.633 & 0.673 & 0.614 & 0.680 & 0.688 & \textbf{0.691} \\
Austronesian & Filipino (fil) & 0.698 & 0.617 & 0.689 & 0.713 & 0.661 & 0.714 & \textbf{0.723} & 0.719 \\
Austronesian & Indonesian (id) & 0.706 & 0.613 & 0.702 & 0.710 & 0.680 & 0.710 & \textbf{0.717} & 0.693 \\
Austronesian & Malay (ms) & \textbf{0.738} & 0.613 & 0.710 & 0.721 & 0.678 & 0.730 & 0.724 & 0.726 \\
\midrule
Romance & Spanish (es) & 0.732 & 0.638 & 0.737 & 0.729 & 0.709 & \textbf{0.738} & 0.727 & 0.732 \\
Romance & Portuguese (pt) & 0.759 & 0.647 & 0.748 & \textbf{0.766} & 0.697 & 0.759 & 0.760 & 0.756 \\
Romance & French (fr) & 0.779 & 0.644 & 0.774 & 0.779 & 0.732 & 0.782 & 0.780 & \textbf{0.787} \\
Romance & Italian (it) & 0.724 & 0.600 & 0.732 & 0.736 & 0.679 & \textbf{0.748} & 0.728 & 0.739 \\
Romance & Romanian (ro) & \textbf{0.746} & 0.597 & 0.739 & 0.738 & 0.661 & 0.738 & 0.744 & 0.738 \\
\bottomrule
\multicolumn{10}{l}{}
\end{tabular}}
\caption{Per-language Belebele accuracy $\uparrow$ (2-shot). D.-Rev.$_{10}$, E.-Rev.$_{10}$, Freeze$_{10}$, and L2-SP$_{10}$ use the targeted layer window $[10,16)$; Expert, E.-Rev.$_{10}$, and the targeted columns report the model trained on the language's family. \textbf{Bold} marks the best value at the displayed precision, including ties.}
\label{tab:belebele_perlang}
\end{table}

\begin{table}[p]
\centering
\scriptsize
\resizebox{\textwidth}{!}{%
\begin{tabular}{llrrrrrrrr}
\toprule
\textbf{Family} & \textbf{Language} & \textbf{Base} & \textbf{Dense} & \textbf{Expert} & \textbf{Soup} & \textbf{D.-Rev.$_5$} & \textbf{E.-Rev.$_5$} & \textbf{Freeze$_5$} & \textbf{L2-SP$_5$} \\
\midrule
Slavic & Croatian (hr) & 23.1 & 55.6 & 45.0 & 44.1 & \textbf{58.3} & 54.6 & 45.0 & 45.3 \\
Slavic & Macedonian (mk) & 27.8 & 57.8 & 49.5 & 40.8 & \textbf{62.5} & 58.3 & 51.2 & 52.0 \\
Slavic & Russian (ru) & 22.2 & 56.7 & 38.2 & 39.4 & \textbf{57.9} & 39.5 & 35.2 & 33.1 \\
Slavic & Slovak (sk) & 25.4 & 55.9 & 47.9 & 49.3 & \textbf{60.3} & 57.1 & 49.1 & 47.2 \\
Slavic & Serbian (sr) & 29.0 & 55.6 & 51.2 & 45.4 & \textbf{59.9} & 57.0 & 51.8 & 50.0 \\
Slavic & Ukrainian (uk) & 28.5 & \underline{58.8} & 47.0 & 46.1 & \textbf{59.8} & 51.3 & 43.8 & 44.4 \\
\midrule
Germanic & Afrikaans (af) & 20.2 & 69.8 & 65.4 & 46.8 & \textbf{71.0} & 61.3 & 62.0 & 64.3 \\
Germanic & Danish (da) & 21.9 & 64.6 & 59.1 & 49.4 & \textbf{66.6} & 64.9 & 61.9 & 62.1 \\
Germanic & Luxembourgish (lb) & 36.0 & 56.7 & 56.5 & 52.7 & \textbf{65.2} & 63.8 & 62.8 & 61.8 \\
Germanic & Dutch (nl) & 25.1 & 54.5 & 50.5 & 39.8 & \textbf{56.1} & \underline{55.9} & 47.3 & 47.4 \\
\midrule
Indic & Bengali (bn) & 21.2 & \textbf{50.7} & 30.2 & 44.2 & 45.0 & 30.8 & 32.9 & 32.6 \\
Indic & Hindi (hi) & 21.3 & \textbf{51.5} & 32.1 & 48.0 & 46.1 & 33.9 & 35.0 & 35.7 \\
Indic & Kannada (kn) & 34.2 & 45.2 & 36.0 & \textbf{49.6} & 40.9 & 34.1 & 34.1 & 34.8 \\
Indic & Malayalam (ml) & 31.7 & \underline{43.8} & 36.8 & \textbf{44.7} & 39.9 & 35.4 & 34.6 & 36.5 \\
Indic & Marathi (mr) & 30.1 & \textbf{48.9} & 35.1 & 47.7 & 43.5 & 34.2 & 34.2 & 35.7 \\
Indic & Nepali (ne) & 24.8 & \textbf{54.2} & 38.5 & 48.9 & 50.2 & 40.8 & 41.1 & 41.5 \\
Indic & Tamil (ta) & 27.9 & 45.4 & 34.9 & \textbf{46.5} & 40.5 & 33.2 & 32.3 & 33.3 \\
Indic & Telugu (te) & 33.0 & 41.7 & 35.4 & \textbf{45.3} & 36.2 & 31.9 & 32.7 & 34.4 \\
\midrule
Austronesian & Cebuano (ceb) & 40.0 & 57.3 & 12.0 & 56.2 & \textbf{62.9} & 9.7 & 10.9 & 10.1 \\
Austronesian & Filipino (fil) & 29.9 & 61.1 & 17.0 & 51.1 & \textbf{63.7} & 10.6 & 13.8 & 14.2 \\
Austronesian & Indonesian (id) & 19.1 & 59.8 & 20.5 & 50.5 & \textbf{62.2} & 13.3 & 17.7 & 17.1 \\
Austronesian & Javanese (jv) & 29.4 & 55.0 & 16.4 & 47.8 & \textbf{57.6} & 12.5 & 18.3 & 17.3 \\
Austronesian & Malay (ms) & 20.1 & 60.6 & 21.5 & 50.9 & \textbf{62.2} & 14.0 & 16.9 & 16.4 \\
Austronesian & Samoan (sm) & 35.5 & \textbf{49.5} & 28.2 & 39.4 & \underline{48.9} & 25.3 & 29.7 & 30.0 \\
\midrule
Romance & Spanish (es) & 28.2 & 55.2 & 42.8 & 36.5 & \textbf{57.6} & 40.4 & 38.3 & 36.6 \\
Romance & French (fr) & 25.4 & 63.6 & 56.1 & 49.9 & \textbf{65.2} & 57.6 & 51.2 & 51.2 \\
Romance & Galician (gl) & 25.4 & 58.0 & 54.1 & 46.6 & \textbf{63.6} & 57.5 & 53.2 & 53.1 \\
Romance & Italian (it) & 27.2 & 55.1 & 48.8 & 36.8 & \textbf{59.3} & 51.6 & 38.8 & 37.7 \\
Romance & Portuguese (pt) & 24.2 & 61.3 & 57.9 & 48.1 & \textbf{66.4} & 61.2 & 49.1 & 49.0 \\
Romance & Romanian (ro) & 30.8 & 58.5 & 60.5 & 51.5 & \textbf{64.2} & 62.0 & 54.3 & 53.7 \\
\bottomrule
\end{tabular}}
\caption{Per-language FLORES-200 ChrF $\uparrow$ (xx$\to$en, 2-shot). D.-Rev.$_5$, E.-Rev.$_5$, Freeze$_5$, and L2-SP$_5$ use the targeted layer window $[5,11)$; Dense is the pipeline-matched control for D.-Rev.$_5$. Expert, E.-Rev.$_5$, and the targeted training columns report the model trained on the language's family. \textbf{Bold} = best per row; \underline{underline} = within 1 ChrF point of best.}
\label{tab:flores_xx_en_perlang}
\end{table}

\begin{table}[p]
\centering
\scriptsize
\resizebox{\textwidth}{!}{%
\begin{tabular}{llrrrrrrrr}
\toprule
\textbf{Family} & \textbf{Language} & \textbf{Base} & \textbf{Dense} & \textbf{Expert} & \textbf{Soup} & \textbf{D.-Rev.$_5$} & \textbf{E.-Rev.$_5$} & \textbf{Freeze$_5$} & \textbf{L2-SP$_5$} \\
\midrule
Slavic & Croatian (hr) & 36.9 & \textbf{53.1} & 50.7 & 48.9 & \underline{52.5} & \underline{52.7} & 51.7 & 51.6 \\
Slavic & Macedonian (mk) & 40.3 & 47.0 & \underline{51.4} & 50.6 & 50.1 & \textbf{52.0} & 50.1 & 49.2 \\
Slavic & Russian (ru) & 43.3 & \underline{51.0} & 46.1 & 47.7 & \textbf{51.4} & 49.8 & 47.0 & 47.1 \\
Slavic & Slovak (sk) & \underline{45.5} & 43.0 & \underline{45.5} & 43.9 & \underline{45.2} & \textbf{45.6} & 44.1 & \underline{45.0} \\
Slavic & Serbian (sr) & 35.2 & \underline{49.0} & \textbf{49.2} & \underline{49.0} & \underline{48.8} & \underline{48.8} & \underline{48.3} & \underline{48.9} \\
Slavic & Ukrainian (uk) & 45.3 & \underline{48.9} & 47.6 & \underline{49.2} & \textbf{49.8} & \underline{49.0} & 48.3 & 48.5 \\
\midrule
Germanic & Afrikaans (af) & 40.9 & \textbf{65.1} & 63.9 & 61.4 & \underline{64.8} & \underline{64.8} & 63.3 & \underline{64.5} \\
Germanic & Danish (da) & 55.5 & 61.6 & 62.1 & 60.5 & 62.7 & \textbf{63.9} & \underline{63.1} & \underline{63.1} \\
Germanic & Luxembourgish (lb) & 39.2 & 47.5 & 45.9 & 42.1 & \textbf{49.8} & 48.1 & 47.1 & 47.2 \\
Germanic & Dutch (nl) & 40.0 & 52.4 & 52.3 & 50.3 & \underline{53.9} & \textbf{54.5} & \underline{54.1} & \underline{54.1} \\
\midrule
Indic & Bengali (bn) & 34.1 & \textbf{42.0} & 38.1 & 39.5 & \underline{41.0} & 38.7 & 36.7 & 37.0 \\
Indic & Hindi (hi) & 27.9 & \textbf{43.6} & 40.6 & \underline{43.3} & 42.2 & 40.1 & 39.9 & 40.8 \\
Indic & Kannada (kn) & 35.8 & 37.5 & 38.4 & 38.0 & \underline{40.4} & 39.0 & \textbf{40.8} & 39.6 \\
Indic & Malayalam (ml) & 33.0 & 33.7 & 35.3 & 34.0 & \underline{36.8} & \underline{36.8} & \textbf{37.4} & 36.2 \\
Indic & Marathi (mr) & 26.8 & \textbf{38.5} & 34.0 & 36.6 & \underline{37.6} & 35.8 & 34.6 & 34.1 \\
Indic & Nepali (ne) & 29.8 & \underline{45.6} & 42.6 & 40.1 & \textbf{46.6} & 44.1 & 42.8 & 42.7 \\
Indic & Tamil (ta) & 33.5 & \textbf{41.1} & 37.4 & 39.5 & \underline{40.1} & 39.4 & 38.8 & 37.9 \\
Indic & Telugu (te) & 34.1 & 31.4 & 32.9 & \textbf{35.5} & 33.3 & 34.0 & 33.6 & 32.6 \\
\midrule
Austronesian & Cebuano (ceb) & 38.9 & 54.3 & 25.3 & 46.7 & \textbf{56.4} & 26.9 & 30.6 & 24.5 \\
Austronesian & Filipino (fil) & 44.2 & \underline{57.0} & 25.6 & 53.6 & \textbf{57.1} & 23.3 & 22.5 & 20.2 \\
Austronesian & Indonesian (id) & 41.4 & \textbf{63.5} & 28.4 & \textbf{63.5} & \textbf{63.5} & 28.2 & 35.8 & 35.3 \\
Austronesian & Javanese (jv) & 37.3 & \textbf{39.9} & 20.9 & 36.9 & 29.8 & 19.3 & 24.5 & 23.0 \\
Austronesian & Malay (ms) & 53.0 & 61.6 & 43.9 & 60.9 & \textbf{62.7} & 43.6 & 36.6 & 34.5 \\
Austronesian & Samoan (sm) & 16.6 & \underline{47.1} & 40.4 & 19.3 & \textbf{47.9} & 39.4 & 38.9 & 38.3 \\
\midrule
Romance & Spanish (es) & 41.3 & \textbf{53.3} & 51.2 & 48.8 & \underline{52.8} & 51.6 & 45.5 & 44.3 \\
Romance & French (fr) & 36.7 & \textbf{65.2} & 63.2 & 61.6 & \textbf{65.2} & \underline{64.7} & 58.4 & 56.2 \\
Romance & Galician (gl) & 37.9 & \underline{56.2} & 54.7 & 48.1 & \textbf{57.0} & \underline{56.5} & 48.5 & 49.1 \\
Romance & Italian (it) & 44.7 & \underline{53.7} & 52.8 & 48.4 & \textbf{54.3} & \textbf{54.3} & 46.6 & 44.9 \\
Romance & Portuguese (pt) & 40.2 & \underline{65.1} & 63.3 & 61.0 & \textbf{65.2} & \underline{65.0} & 58.1 & 56.1 \\
Romance & Romanian (ro) & 51.2 & 49.2 & \underline{53.2} & \underline{53.5} & 52.1 & \textbf{54.2} & 51.5 & 50.3 \\
\bottomrule
\end{tabular}}
\caption{Per-language FLORES-200 ChrF $\uparrow$ (en$\to$xx, 2-shot). D.-Rev.$_5$, E.-Rev.$_5$, Freeze$_5$, and L2-SP$_5$ use the targeted layer window $[5,11)$; Dense is the pipeline-matched control for D.-Rev.$_5$. Expert, E.-Rev.$_5$, and the targeted training columns report the model trained on the language's family. \textbf{Bold} = best per row; \underline{underline} = within 1 ChrF point of best.}
\label{tab:flores_en_xx_perlang}
\end{table}

\clearpage

\subsection{Per-Language Held-Out Results Tables}
\label{app:heldout_perlang}

Tables~\ref{tab:heldout_ppl_perlang}--\ref{tab:heldout_flores_en_xx_perlang} report results for the same 22 languages excluded from CPT, complementing the family-averaged perplexity and Belebele results in Tables~\ref{tab:perplexity_heldout} and~\ref{tab:belebele_targeted_heldout}.
Perplexity covers both targeted windows; Belebele uses $[10,16)$ and FLORES uses $[5,11)$. For each language, Expert, Freeze, L2-SP, and E.-Rev.\ report the model trained on its family; Base, Dense, D.-Rev., and Soup are shared models. Reverted models restore the indicated layers to Base weights. Family summaries average languages equally (and both directions for FLORES); overall summaries weight the five families equally.

\begin{table}[p]
\centering
\scriptsize
\setlength{\tabcolsep}{3pt}
\resizebox{\textwidth}{!}{%
\begin{tabular}{llrrrrrrrrrrrr}
\toprule
& & \multicolumn{4}{c}{\textbf{Shared baselines}} & \multicolumn{4}{c}{$\mathbf{[5,11)}$} & \multicolumn{4}{c}{$\mathbf{[10,16)}$} \\
\cmidrule(lr){3-6} \cmidrule(lr){7-10} \cmidrule(lr){11-14}
\textbf{Family} & \textbf{Language} & \textbf{Base} & \textbf{Dense} & \textbf{Expert} & \textbf{Soup} & \textbf{D.-Rev.} & \textbf{E.-Rev.} & \textbf{Freeze} & \textbf{L2-SP} & \textbf{D.-Rev.} & \textbf{E.-Rev.} & \textbf{Freeze} & \textbf{L2-SP} \\
\midrule
Slavic & Bulgarian (\texttt{bg}) & 6.46 & 8.13 & 7.21 & \textbf{6.23} & 8.87 & 7.51 & 7.12 & 7.14 & 8.07 & 7.09 & 7.14 & 7.15 \\
Slavic & Czech (\texttt{cs}) & 7.90 & 8.48 & 7.75 & \textbf{7.56} & 8.73 & 7.88 & 7.75 & 7.79 & 8.52 & 7.78 & 7.75 & 7.82 \\
Slavic & Lithuanian (\texttt{lt}) & 7.38 & 9.57 & 7.56 & \textbf{7.17} & 9.52 & 7.51 & 7.58 & 7.60 & 9.56 & 7.53 & 7.60 & 7.63 \\
Slavic & Latvian (\texttt{lv}) & 6.43 & 8.45 & 6.67 & \textbf{6.29} & 8.40 & 6.62 & 6.69 & 6.70 & 8.41 & 6.64 & 6.70 & 6.73 \\
Slavic & Polish (\texttt{pl}) & 7.87 & 9.20 & 7.82 & \textbf{7.51} & 9.21 & 7.80 & 7.88 & 7.91 & 9.24 & 7.86 & 7.89 & 7.95 \\
Slavic & Slovenian (\texttt{sl}) & 8.09 & 11.13 & 8.85 & \textbf{7.82} & 10.80 & 8.67 & 8.74 & 8.79 & 10.92 & 8.65 & 8.69 & 8.79 \\
\midrule
Germanic & Icelandic (\texttt{is}) & \textbf{8.05} & 14.68 & 10.30 & 8.11 & 14.18 & 10.20 & 10.04 & 10.00 & 14.18 & 9.80 & 10.07 & 10.01 \\
Germanic & Norwegian (\texttt{no}) & 8.43 & 9.41 & 8.35 & \textbf{8.03} & 9.37 & 8.39 & 8.38 & 8.41 & 9.52 & 8.38 & 8.38 & 8.37 \\
Germanic & Swedish (\texttt{sv}) & 9.20 & 10.24 & 9.04 & \textbf{8.74} & 10.24 & 9.09 & 9.09 & 9.12 & 10.29 & 9.08 & 9.08 & 9.07 \\
\midrule
Indic & Assamese (\texttt{as}) & \textbf{10.28} & 14.91 & 11.25 & 10.43 & 13.83 & 10.93 & 11.17 & 11.14 & 13.59 & 10.70 & 11.20 & 11.10 \\
Indic & Gujarati (\texttt{gu}) & 7.30 & 10.97 & 7.37 & \textbf{7.04} & 11.08 & 7.47 & 7.61 & 7.60 & 11.24 & 7.48 & 7.62 & 7.50 \\
Indic & Odia (\texttt{or}) & 4.62 & 5.45 & 4.86 & \textbf{4.56} & 5.43 & 4.86 & 4.81 & 4.86 & 5.43 & 4.79 & 4.88 & 4.85 \\
Indic & Punjabi (\texttt{pa}) & 5.12 & 6.32 & 5.31 & \textbf{5.01} & 6.29 & 5.31 & 5.40 & 5.40 & 6.29 & 5.28 & 5.41 & 5.36 \\
Indic & Sindhi (\texttt{sd}) & \textbf{12.87} & 26.61 & 14.94 & 13.20 & 26.66 & 15.04 & 14.50 & 14.58 & 24.11 & 14.68 & 14.70 & 14.58 \\
Indic & Sinhala (\texttt{si}) & \textbf{7.47} & 16.03 & 9.01 & \textbf{7.47} & 16.21 & 9.04 & 9.15 & 9.12 & 15.81 & 8.87 & 9.48 & 9.23 \\
Indic & Urdu (\texttt{ur}) & 15.29 & 22.61 & 15.55 & \textbf{14.87} & 23.07 & 15.88 & 15.79 & 15.83 & 23.36 & 15.79 & 15.76 & 15.68 \\
\midrule
Austronesian & Ilocano (\texttt{ilo}) & 25.41 & 40.41 & 35.42 & \textbf{25.13} & 41.45 & 34.34 & 36.47 & 36.18 & 40.38 & 32.44 & 33.99 & 34.45 \\
Austronesian & Malagasy (\texttt{mg}) & \textbf{9.97} & 26.89 & 28.91 & 10.29 & 29.83 & 30.06 & 30.31 & 30.49 & 24.63 & 20.68 & 27.02 & 26.70 \\
Austronesian & M\={a}ori (\texttt{mi}) & \textbf{14.10} & 36.44 & 38.41 & 14.45 & 39.50 & 44.93 & 40.36 & 41.04 & 39.44 & 36.63 & 38.46 & 38.56 \\
Austronesian & Sundanese (\texttt{su}) & \textbf{17.15} & 34.05 & 35.29 & 17.57 & 34.87 & 37.00 & 39.18 & 39.94 & 35.33 & 32.01 & 35.32 & 34.54 \\
Austronesian & Waray (\texttt{war}) & 3.57 & 4.01 & 3.72 & \textbf{3.39} & 4.01 & 3.66 & 3.61 & 3.61 & 3.96 & 3.64 & 3.63 & 3.63 \\
\midrule
Romance & Catalan (\texttt{ca}) & 7.90 & 9.07 & 7.93 & \textbf{7.57} & 9.31 & 8.01 & 7.93 & 7.89 & 9.21 & 7.94 & 7.97 & 7.97 \\
\bottomrule
\end{tabular}}
\caption{Per-language held-out validation perplexity $\downarrow$ for both targeted windows. D.-Rev.\ and E.-Rev.\ restore the indicated layers to Base after Dense and Expert CPT, respectively; Freeze and L2-SP preserve those layers during training. Expert, Freeze, L2-SP, and E.-Rev.\ use the model trained on the language's family; Base, Dense, D.-Rev., and Soup are shared models. \textbf{Bold} marks the best value at the displayed precision, including ties.}
\label{tab:heldout_ppl_perlang}
\end{table}

\begin{table}[p]
\centering
\scriptsize
\setlength{\tabcolsep}{3pt}
\resizebox{\textwidth}{!}{%
\begin{tabular}{llrrrrrrrr}
\toprule
& & \multicolumn{4}{c}{\textbf{Shared baselines}} & \multicolumn{4}{c}{$\mathbf{[10,16)}$} \\
\cmidrule(lr){3-6} \cmidrule(lr){7-10}
\textbf{Family} & \textbf{Language} & \textbf{Base} & \textbf{Dense} & \textbf{Expert} & \textbf{Soup} & \textbf{D.-Rev.} & \textbf{E.-Rev.} & \textbf{Freeze} & \textbf{L2-SP} \\
\midrule
Slavic & Bulgarian (\texttt{bg}) & 0.707 & 0.629 & 0.714 & 0.718 & 0.676 & \textbf{0.730} & 0.724 & 0.726 \\
Slavic & Czech (\texttt{cs}) & \textbf{0.756} & 0.628 & 0.734 & 0.748 & 0.683 & 0.746 & 0.730 & 0.736 \\
Slavic & Lithuanian (\texttt{lt}) & \textbf{0.724} & 0.614 & 0.721 & 0.716 & 0.668 & 0.723 & 0.718 & 0.710 \\
Slavic & Latvian (\texttt{lv}) & \textbf{0.732} & 0.576 & 0.699 & 0.707 & 0.636 & 0.714 & 0.696 & 0.712 \\
Slavic & Polish (\texttt{pl}) & 0.707 & 0.570 & 0.692 & 0.709 & 0.644 & 0.707 & 0.692 & \textbf{0.712} \\
Slavic & Slovenian (\texttt{sl}) & \textbf{0.723} & 0.569 & 0.699 & 0.716 & 0.637 & 0.708 & 0.688 & 0.702 \\
\midrule
Germanic & Icelandic (\texttt{is}) & \textbf{0.664} & 0.524 & 0.636 & 0.656 & 0.554 & 0.649 & 0.639 & 0.644 \\
Germanic & Norwegian (\texttt{no}) & \textbf{0.743} & 0.613 & 0.714 & 0.737 & 0.697 & 0.731 & \textbf{0.743} & 0.732 \\
Germanic & Swedish (\texttt{sv}) & \textbf{0.759} & 0.638 & 0.748 & 0.756 & 0.706 & 0.743 & 0.756 & 0.751 \\
\midrule
Indic & Assamese (\texttt{as}) & \textbf{0.560} & 0.460 & 0.539 & 0.548 & 0.516 & 0.549 & 0.552 & 0.550 \\
Indic & Gujarati (\texttt{gu}) & \textbf{0.593} & 0.500 & 0.578 & 0.571 & 0.551 & 0.588 & 0.584 & 0.590 \\
Indic & Odia (\texttt{or}) & \textbf{0.542} & 0.430 & 0.519 & 0.536 & 0.461 & 0.514 & 0.522 & 0.520 \\
Indic & Punjabi (\texttt{pa}) & 0.618 & 0.500 & 0.598 & \textbf{0.626} & 0.559 & 0.621 & 0.608 & 0.609 \\
Indic & Sindhi (\texttt{sd}) & 0.478 & 0.421 & 0.473 & \textbf{0.501} & 0.434 & 0.457 & 0.462 & 0.462 \\
Indic & Sinhala (\texttt{si}) & \textbf{0.607} & 0.470 & 0.586 & 0.604 & 0.548 & 0.606 & 0.602 & 0.604 \\
Indic & Urdu (\texttt{ur}) & \textbf{0.637} & 0.551 & 0.609 & 0.624 & 0.579 & 0.630 & 0.619 & 0.627 \\
\midrule
Austronesian & Ilocano (\texttt{ilo}) & \textbf{0.410} & 0.349 & 0.384 & 0.398 & 0.348 & 0.382 & 0.407 & 0.387 \\
Austronesian & Malagasy (\texttt{mg}) & \textbf{0.529} & 0.391 & 0.481 & 0.527 & 0.409 & 0.494 & 0.506 & 0.499 \\
Austronesian & M\={a}ori (\texttt{mi}) & \textbf{0.367} & 0.317 & 0.304 & 0.351 & 0.321 & 0.308 & 0.319 & 0.323 \\
Austronesian & Sundanese (\texttt{su}) & \textbf{0.578} & 0.427 & 0.527 & 0.570 & 0.451 & 0.524 & 0.558 & 0.556 \\
Austronesian & Waray (\texttt{war}) & \textbf{0.550} & 0.431 & 0.462 & 0.530 & 0.464 & 0.493 & 0.493 & 0.510 \\
\midrule
Romance & Catalan (\texttt{ca}) & \textbf{0.752} & 0.600 & 0.731 & \textbf{0.752} & 0.682 & 0.744 & 0.741 & 0.744 \\
\bottomrule
\end{tabular}}
\caption{Per-language held-out Belebele accuracy $\uparrow$ (2-shot), with all targeted interventions at $[10,16)$. Expert, Freeze, L2-SP, and E.-Rev.\ use the model trained on the language's family; Base, Dense, D.-Rev., and Soup are shared models. \textbf{Bold} marks the best value at the displayed precision, including ties.}
\label{tab:heldout_belebele_perlang}
\end{table}

\begin{table}[p]
\centering
\scriptsize
\setlength{\tabcolsep}{3pt}
\resizebox{\textwidth}{!}{%
\begin{tabular}{llrrrrrrrr}
\toprule
& & \multicolumn{4}{c}{\textbf{Shared baselines}} & \multicolumn{4}{c}{$\mathbf{[5,11)}$} \\
\cmidrule(lr){3-6} \cmidrule(lr){7-10}
\textbf{Family} & \textbf{Language} & \textbf{Base} & \textbf{Dense} & \textbf{Expert} & \textbf{Soup} & \textbf{D.-Rev.} & \textbf{E.-Rev.} & \textbf{Freeze} & \textbf{L2-SP} \\
\midrule
Slavic & Bulgarian (\texttt{bg}) & 25.6 & 56.9 & 48.4 & 43.8 & \textbf{62.0} & 56.8 & 47.1 & 48.6 \\
Slavic & Czech (\texttt{cs}) & 25.3 & 57.7 & 49.5 & 49.5 & \textbf{59.8} & 57.6 & 49.1 & 49.3 \\
Slavic & Lithuanian (\texttt{lt}) & 28.9 & 50.6 & 44.7 & 45.1 & 52.0 & \textbf{52.5} & 45.1 & 48.0 \\
Slavic & Latvian (\texttt{lv}) & 37.6 & \textbf{57.5} & 50.2 & 52.0 & 56.6 & 56.6 & 51.7 & 51.9 \\
Slavic & Polish (\texttt{pl}) & 24.8 & 51.7 & 38.6 & 40.8 & \textbf{54.4} & 46.0 & 36.1 & 35.8 \\
Slavic & Slovenian (\texttt{sl}) & 25.9 & 55.4 & 43.6 & 44.5 & \textbf{56.5} & 54.1 & 46.6 & 47.6 \\
\midrule
Germanic & Icelandic (\texttt{is}) & 42.4 & 50.6 & 51.7 & 49.0 & 52.2 & \textbf{54.9} & 53.4 & 53.5 \\
Germanic & Norwegian (\texttt{no}) & 24.0 & 59.8 & 46.7 & 40.6 & \textbf{61.8} & 60.4 & 58.0 & 56.6 \\
Germanic & Swedish (\texttt{sv}) & 21.9 & 61.2 & 57.0 & 48.5 & 63.0 & \textbf{63.1} & 59.1 & 60.2 \\
\midrule
Indic & Assamese (\texttt{as}) & 33.5 & 40.1 & 31.0 & \textbf{41.6} & 41.2 & 31.2 & 32.5 & 33.8 \\
Indic & Gujarati (\texttt{gu}) & 28.8 & \textbf{49.9} & 35.2 & 47.4 & 49.3 & 35.5 & 37.7 & 39.6 \\
Indic & Odia (\texttt{or}) & 44.4 & 40.9 & 37.0 & \textbf{44.9} & 38.3 & 37.8 & 35.4 & 34.3 \\
Indic & Punjabi (\texttt{pa}) & 38.4 & \textbf{50.0} & 38.8 & 47.4 & 49.1 & 39.6 & 42.0 & 44.0 \\
Indic & Sindhi (\texttt{sd}) & 34.8 & 45.6 & 34.6 & 45.3 & \textbf{46.8} & 37.7 & 39.2 & 39.1 \\
Indic & Sinhala (\texttt{si}) & 38.8 & 39.9 & 36.0 & \textbf{44.6} & 37.2 & 35.8 & 34.0 & 36.2 \\
Indic & Urdu (\texttt{ur}) & 28.6 & \textbf{50.7} & 32.4 & 45.2 & 49.9 & 34.7 & 33.8 & 36.1 \\
\midrule
Austronesian & Ilocano (\texttt{ilo}) & 40.1 & 36.9 & 28.8 & 38.6 & \textbf{41.1} & 29.6 & 30.2 & 29.0 \\
Austronesian & Malagasy (\texttt{mg}) & 35.0 & 31.1 & 24.8 & \textbf{38.2} & 34.5 & 26.3 & 29.8 & 29.3 \\
Austronesian & M\={a}ori (\texttt{mi}) & 28.0 & 24.9 & 21.9 & \textbf{30.9} & 28.8 & 23.2 & 23.1 & 22.3 \\
Austronesian & Sundanese (\texttt{su}) & 31.2 & 48.9 & 20.8 & 48.1 & \textbf{51.7} & 18.7 & 23.1 & 21.1 \\
Austronesian & Waray (\texttt{war}) & 41.1 & 53.7 & 19.3 & 53.9 & \textbf{59.3} & 15.3 & 20.6 & 19.8 \\
\midrule
Romance & Catalan (\texttt{ca}) & 31.0 & 60.1 & 50.5 & 36.4 & \textbf{64.8} & 51.2 & 44.5 & 44.5 \\
\bottomrule
\end{tabular}}
\caption{Per-language held-out FLORES-200 ChrF (xx$\to$en) $\uparrow$, with all targeted interventions at $[5,11)$ and the postprocessing in Appendix~\ref{app:flores_eval_protocol}. Expert, Freeze, L2-SP, and E.-Rev.\ use the model trained on the language's family; Base, Dense, D.-Rev., and Soup are shared models. \textbf{Bold} marks the best value at the displayed precision, including ties.}
\label{tab:heldout_flores_xx_en_perlang}
\end{table}

\begin{table}[p]
\centering
\scriptsize
\setlength{\tabcolsep}{3pt}
\resizebox{\textwidth}{!}{%
\begin{tabular}{llrrrrrrrr}
\toprule
& & \multicolumn{4}{c}{\textbf{Shared baselines}} & \multicolumn{4}{c}{$\mathbf{[5,11)}$} \\
\cmidrule(lr){3-6} \cmidrule(lr){7-10}
\textbf{Family} & \textbf{Language} & \textbf{Base} & \textbf{Dense} & \textbf{Expert} & \textbf{Soup} & \textbf{D.-Rev.} & \textbf{E.-Rev.} & \textbf{Freeze} & \textbf{L2-SP} \\
\midrule
Slavic & Bulgarian (\texttt{bg}) & 35.7 & 43.1 & 48.5 & \textbf{51.3} & 40.3 & 47.1 & 48.6 & 49.0 \\
Slavic & Czech (\texttt{cs}) & 36.7 & 42.9 & 43.3 & 41.7 & 41.5 & \textbf{44.7} & 43.1 & 43.7 \\
Slavic & Lithuanian (\texttt{lt}) & 25.9 & 34.9 & \textbf{41.4} & 40.9 & 34.8 & 41.1 & 39.3 & 38.9 \\
Slavic & Latvian (\texttt{lv}) & 43.4 & 32.5 & 41.7 & \textbf{45.1} & 34.1 & 43.5 & 40.0 & 41.6 \\
Slavic & Polish (\texttt{pl}) & 35.2 & 34.0 & 35.5 & \textbf{39.3} & 32.1 & 38.0 & 36.4 & 37.0 \\
Slavic & Slovenian (\texttt{sl}) & 33.4 & 38.8 & 42.4 & 42.9 & 39.6 & \textbf{44.7} & 42.2 & 42.2 \\
\midrule
Germanic & Icelandic (\texttt{is}) & 32.0 & 18.9 & 27.8 & \textbf{34.9} & 19.6 & 30.4 & 27.4 & 28.6 \\
Germanic & Norwegian (\texttt{no}) & 40.7 & 51.1 & 47.5 & 43.6 & 52.5 & \textbf{54.2} & 52.9 & 53.2 \\
Germanic & Swedish (\texttt{sv}) & 46.2 & 59.0 & 59.9 & 60.6 & 58.3 & \textbf{63.2} & 60.7 & 61.0 \\
\midrule
Indic & Assamese (\texttt{as}) & 15.9 & 12.0 & 17.9 & 17.9 & 13.5 & 17.6 & 18.6 & \textbf{18.7} \\
Indic & Gujarati (\texttt{gu}) & 28.9 & 14.7 & 25.7 & \textbf{34.7} & 15.2 & 25.8 & 24.0 & 23.6 \\
Indic & Odia (\texttt{or}) & \textbf{19.1} & 8.1 & 14.5 & 17.8 & 9.3 & 14.1 & 13.3 & 12.2 \\
Indic & Punjabi (\texttt{pa}) & 33.5 & 19.5 & 27.3 & \textbf{34.0} & 19.4 & 28.5 & 24.7 & 25.6 \\
Indic & Sindhi (\texttt{sd}) & \textbf{16.4} & 4.5 & 9.2 & 14.0 & 3.4 & 8.0 & 9.3 & 9.0 \\
Indic & Sinhala (\texttt{si}) & \textbf{25.4} & 3.0 & 7.8 & 24.3 & 2.8 & 7.6 & 6.5 & 6.3 \\
Indic & Urdu (\texttt{ur}) & 26.5 & 19.0 & 27.2 & \textbf{33.3} & 19.5 & 29.0 & 26.1 & 26.6 \\
\midrule
Austronesian & Ilocano (\texttt{ilo}) & 18.1 & 21.3 & 9.4 & 14.0 & \textbf{26.6} & 9.7 & 9.9 & 9.0 \\
Austronesian & Malagasy (\texttt{mg}) & \textbf{23.1} & 9.5 & 9.6 & 14.8 & 9.6 & 10.7 & 10.9 & 11.1 \\
Austronesian & M\={a}ori (\texttt{mi}) & 16.9 & 15.0 & 17.4 & 13.8 & \textbf{18.0} & 17.7 & 17.5 & 17.6 \\
Austronesian & Sundanese (\texttt{su}) & \textbf{32.3} & 29.4 & 15.6 & 29.3 & 29.3 & 15.4 & 15.7 & 15.6 \\
Austronesian & Waray (\texttt{war}) & 30.9 & 43.6 & 17.4 & 37.1 & \textbf{43.7} & 15.9 & 13.2 & 11.0 \\
\midrule
Romance & Catalan (\texttt{ca}) & 49.3 & 56.9 & 58.5 & 55.3 & 57.8 & \textbf{59.3} & 54.1 & 53.6 \\
\bottomrule
\end{tabular}}
\caption{Per-language held-out FLORES-200 ChrF (en$\to$xx) $\uparrow$, with all targeted interventions at $[5,11)$ and the postprocessing in Appendix~\ref{app:flores_eval_protocol}. Expert, Freeze, L2-SP, and E.-Rev.\ use the model trained on the language's family; Base, Dense, D.-Rev., and Soup are shared models. \textbf{Bold} marks the best value at the displayed precision, including ties.}
\label{tab:heldout_flores_en_xx_perlang}
\end{table}

\end{document}